\documentclass{article}

\PassOptionsToPackage{numbers,compress}{natbib}
\usepackage[preprint]{neurips_2026}

\usepackage[utf8]{inputenc} 
\usepackage[T1]{fontenc}    
\usepackage[hidelinks]{hyperref}       
\usepackage{url}            
\usepackage{longtable}
\usepackage{booktabs}       
\usepackage{amsfonts}       
\usepackage{nicefrac}       
\usepackage{microtype}      
\usepackage[table]{xcolor}         
\usepackage{amsmath, amsthm, amssymb}
\usepackage{bm}
\usepackage{graphicx, multirow}
\usepackage{wrapfig}
\theoremstyle{plain}
\newtheorem{theorem}{Theorem}[section]

\newtheorem{corollary}[theorem]{Corollary}
\theoremstyle{definition}

\theoremstyle{remark}
\newtheorem{remark}[theorem]{Remark}

\DeclareMathOperator*{\argmin}{arg\,min}
\newcommand{\norm}[1]{\left\lVert#1\right\rVert}

\usepackage{caption}
\title{Principled Design of Diffusion-based Optimizers for Inverse Problems}

\author{%
  Julio Oscanoa\thanks{Equal contribution.} \\
  Bioengineering, Radiology\\
  Stanford University\\
  \texttt{joscanoa@stanford.edu} \\
  \And
  Irmak Sivgin$^{*}$ \\
Electrical Engineering \\
  Stanford University \\
  \texttt{isivgin@stanford.edu} \\
\And
  Cagan Alkan$^{*}$ \\
  Electrical Engineering \\
  Stanford University \\
  \texttt{calkan@stanford.edu} \\
\And
  Daniel Ennis \\
  Radiology \\
  Stanford University \\
  \texttt{dbe@stanford.edu} \\
\And
  John Pauly \\
  Electrical Engineering \\
  Stanford University \\
  \texttt{pauly@stanford.edu} \\
  \And
  Mert Pilanci\thanks{Equal senior contribution.} \\
  Electrical Engineering \\
  Stanford University \\
  \texttt{pilanci@stanford.edu} \\
  \And
  Shreyas Vasanawala$^{\dagger}$ \\
  Radiology \\
  Stanford University \\
  \texttt{vasanawala@stanford.edu} \\
}

\begin{document}

\maketitle

\begin{abstract}
  Score-based diffusion models achieve state-of-the-art performance for inverse problems, but their practical deployment is hindered by long inference times and cumbersome hyperparameter tuning. While pretrained diffusion models can be reused across tasks without retraining, inference-time hyperparameters such as the noise schedule and posterior sampling weights typically require ad-hoc adjustment for each problem setup.
  We propose principled reparameterizations that induce invariances, allowing the same hyperparameters to be reused across multiple problems without re-tuning.
  In addition, building on the RED-diff framework, which reformulates posterior sampling as an optimization problem, we further develop the OptDiff pipeline. OptDiff provides a simplified tuning framework that facilitates the integration of convex optimization tools to accelerate inference.
  Experiments on image reconstruction, deblurring, and super-resolution show substantial speedups and improved image quality. Our code is available at \href{https://github.com/MRSRL/OptDiff}{https://github.com/MRSRL/OptDiff}.
\end{abstract}

\section{Introduction}

The flexibility and expressiveness of diffusion models \cite{song2021denoising, ho2020denoising} as powerful image priors have made them strong candidates for solving inverse problems, where the goal is to reconstruct data from incomplete and noisy measurements \cite{jalal2021robust, chung2022diffusion, mardani2023variational}. The key idea is to leverage a pretrained diffusion model to sample from the posterior distribution. However, existing diffusion-based inverse solvers face two major challenges: slow convergence and  strong dependence on hyperparameter selection. The iterative sampling process often requires hundreds of steps, resulting in high computational cost that limits practical deployment in time-sensitive applications. Moreover, reconstruction quality is highly sensitive to hyperparameter choices, and poor selection can lead to instability or degraded performance \cite{zheng2025inversebench, zhang2025improving}.

In this work, we address these challenges through two complementary contributions. First, inspired by maximal update parameterization ($\mu$P) \cite{yang2020feature}, which stabilizes \emph{training} hyperparameters for large neural networks, we introduce principled reparameterizations that stabilize \emph{inference-time} hyperparameters in diffusion-based inverse solvers. These reparameterizations induce invariances across problem configurations, substantially reducing the need for manual tuning. Second, building on the RED-diff framework \cite{mardani2023variational}, which reformulates posterior sampling as a stochastic optimization problem, we propose the \textit{Optimizers with Diffusion-based priors} (OptDiff) pipeline for robust hyperparameter selection. This framework enables the integration of established tools from the optimization literature to design faster inverse solvers. We show that the resulting methods outperform state-of-the-art diffusion-based approaches in both reconstruction quality and inference speed.

\begin{figure*}[t]
  \centering
  \includegraphics[width=0.85\linewidth]{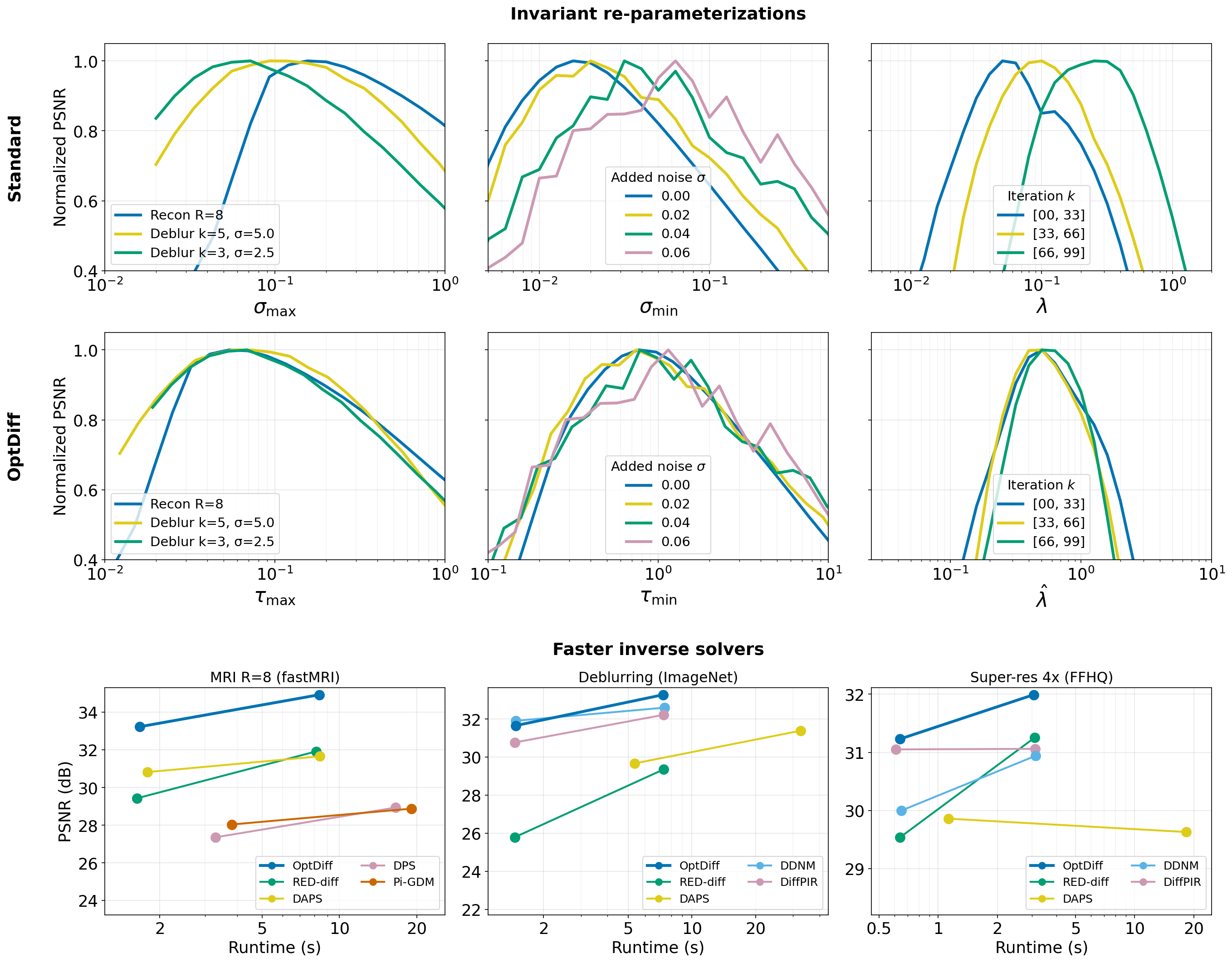}
  \caption[Invariant reparameterizations and faster solvers]{\textbf{(Top)} Comparison of standard and proposed hyperparameters for different experimental settings. Our proposed reparameterization ($\tau_{\max}, \tau_{\min})$ for noise schedule hyperparameters ($\sigma_{\max}, \sigma_{\min})$ produces aligned normalized PSNR curves across different inverse problem settings, showcasing invariance of OptDiff parameters. Similarly, our reparameterized sampler hyperparameter $\hat{\lambda}$ yields aligned curves for different iteration intervals, implying that a single $\hat{\lambda}$ value suffices during the entire reverse process in OptDiff. \textbf{(Bottom)}  Performance comparisons between OptDiff and other diffusion-based inverse problem solvers. Our method OptDiff achieves the best runtime-PSNR tradeoff.
  } \label{fig:invariance}
\end{figure*}

\subsection{Contributions}    
\begin{itemize}
    \item We introduce principled reparameterizations of inference-time hyperparameters for diffusion-based inverse solvers. We organize these into two families: (i) \emph{noise schedule hyperparameters}, such as the start and stop noise levels $\sigma_{\max}$ and $\sigma_{\min}$, and (ii) \emph{sampler hyperparameters}, such as the step size $\alpha$ and regularization weight $\lambda$. The proposed reparameterizations induce invariances across problem setups, enabling robust hyperparameter reuse and substantially reducing the need for manual tuning.

    \item We develop a principled, theory-driven framework for noise schedule design in inverse problems that is solver-agnostic by construction. To the best of our knowledge, this is the first work to study noise schedule design in this context. By analyzing reverse diffusion through the lens of spectral auto-regression \cite{dieleman2024spectral, falck2025fourier}, we derive tolerance-based criteria for selecting $\sigma_{\max}$ and $\sigma_{\min}$ that adapt naturally to the spectral properties of the data and the measurement model. We further demonsrrate empirically that the resulting noise schedule transfer across representative diffusion-based inverse solvers.

    \item For the sampler hyperparameters, we focus on the family of optimization-based solvers in \citet{daras2024survey}. We present a common-descent analysis of the composite data-consistency and diffusion-prior objective that identifies when a single trade-off scalar can be used throughout the optimization, eliminating step-dependent tuning. We develop and validate this framework through RED-diff, whose explicit objective and update rules enable direct analysis.

    \item Building on the RED-diff formulation, we propose the OptDiff pipeline for simplified and robust hyperparameter selection. OptDiff combines invariant reparameterizations with a greedy-optimal hyperparameter scheme, allowing the noise schedule and sampler hyperparameters to be decoupled and tuned efficiently using a small tuning set.

    \item Using the OptDiff pipeline, we demonstrate that incorporating advanced convex optimization techniques leads to significantly faster inference while maintaining or improving reconstruction accuracy. Our results show consistent reductions in both the number of function evaluations and overall runtime.

    \item We evaluate OptDiff across a range of inverse problems, including MRI reconstruction, deblurring, and super-resolution. Across all tasks, our approach achieves state-of-the-art runtime performance while maintaining competitive or improved reconstruction quality.
\end{itemize}

\section{Background}
\label{sec:background}
\subsection{Score-based diffusion models}

Score-based diffusion models involve forward and reverse diffusion processes that are formulated by stochastic differential equations (SDEs). In particular, the forward process for $t\in[0,T]$ is represented by \cite{song2020score}
\begin{equation}
    \mathrm{d} \bm{x}_t=\bm{b}\left(\bm{x}_t, t\right) \,\mathrm{d} t+\sqrt{a(t)}\, \mathrm{d} \bm{W}_t, \label{eq:forward_sde}    
\end{equation}
where $\bm{x}_0\sim p_{\mathrm{data}}$, $\bm{x}_0\in\mathbb{R}^n$, and $\bm{W}_t$ is an $n$-dimensional Wiener process. For sufficiently large $T$, the distribution $p_T(\bm{x}_T)$ approaches an isotropic Gaussian distribution. The reverse diffusion process recovers the data distribution $p(\bm{x}_0)$ from pure noise via the corresponding reverse SDE \cite{anderson1982reverse}
\begin{align}
    \mathrm{d} \bm{x}_t= \left( \bm{b}\left(\bm{x}_t, t\right) - a(t) \nabla_{\bm{x}_t}\log p_t(\bm{x}_t) \right) \mathrm{d} t+ \sqrt{a(t)} \, \mathrm{d} \bar{\bm{W}}_t. \label{eq:reverse_sde}
\end{align}
where $\nabla_{\bm{x}_t}\log p_t(\bm{x}_t)$ is the score function. In practice, the score function is approximated by a neural network $\bm{\epsilon}_{\bm{\theta}}(\bm{x}_t; t)$ which is typically trained using denoising score matching \cite{vincent2011connection}:
\begin{align}
    \argmin_{\bm{\theta}} \mathbb{E}_{t,\bm{x}_0,\bm{x}_t}\left[\norm{\frac{-\bm{\epsilon}_{\bm{\theta}}(\bm{x}_t;t)}{\sigma_t} - \nabla_{\bm{x}_t}\log p_t(\bm{x}_t | \bm{x}_0)}_2^2\right]. \label{eq:dsm}
\end{align}
where $\sigma_t$ is the standard deviation of the perturbation kernel $p_t(\bm{x}_t|\bm{x}_0).$
\paragraph{Noise Schedule.} The reverse process is typically discretized on a decreasing list of noise levels \mbox{$\sigma_{\max}=\sigma_{t_N}>\dots>\sigma_{t_1}=\sigma_{\min}$}, where the start level $\sigma_{\max}$ and stop level $\sigma_{\min}$ are inference-time hyperparameters.
\subsection{Diffusion models for inverse problems}

Inverse problems are described according to a forward model
    $\bm{y} = \mathcal{A}(\bm{x}_0) + \bm{\eta}$, 
where $\bm{y}\in\mathbb{R}^m$ represents the measurements, $\bm{x}_0\in\mathbb{R}^n$ is the signal to be estimated, $\bm{\eta}\sim\mathcal{N}(\mathbf{0}, \sigma^2_{\bm{\eta}} \mathbf{I})$ is the measurement noise, and $\mathcal{A}: \mathbb{R}^n \rightarrow \mathbb{R}^m$ is the measurement operator.
The goal in inverse problems is to recover the underlying signal $\bm{x}_0$ from a potentially incomplete set of measurements $\bm{y}$.

Diffusion-based inverse problem solvers aim to represent and sample from $p(\bm{x}_0|\bm{y})$ in a computationally efficient manner. Given the measurements $\bm{y}$, the conditional score function can be expressed using the Bayes' Theorem as
\begin{align}
    \nabla_{\bm{x}_t}\log p_t(\bm{x}_t|\bm{y}) = \nabla_{\bm{x}_t}\log p_t(\bm{x}_t) + \nabla_{\bm{x}_t}\log p_t(\bm{y} | \bm{x}_t).
\end{align}
Then, the reverse process in \eqref{eq:reverse_sde} can be modified to yield samples from $p(\bm{x}_0|\bm{y})$ \cite{song2022solving}.
The inverse problem scenario additionally requires evaluating the intractable likelihood score function $\nabla_{\bm{x}_t}\log p_t(\bm{y} | \bm{x}_t)$ \cite{chung2022diffusion}. To address this challenge, diffusion-based inverse problem solvers seek to approximate the likelihood score function efficiently using various strategies. 
A comprehensive overview of these methods is provided in the survey by \citet{daras2024survey}.

\subsection{Regularization by denoising diffusion (RED-diff)}

\citet{mardani2023variational} propose a variational inference driven formulation called Regularization by Denoising Diffusion (RED-diff) for solving inverse problems, converting diffusion sampling into an optimization problem. RED-diff represents the complex posterior distribution $p(\bm{x}_0|\bm{y})$ with a variational distribution $q\sim\mathcal{N}(\bm{\mu}, \sigma^2\mathbf{I})$, and minimizes the KL divergence between them. The final optimization objective is:
\begin{equation}
    \min_{\bm{\mu}} \|\bm{y}-\mathcal{A}(\bm{\mu})\|_2^2 +\mathbb{E}_{t, \bm{\epsilon}}\left[\lambda_t\left\|\bm{\epsilon}_{\bm{\theta}}\left(\bm{x}_t ; t\right)-\bm{\epsilon}\right\|_2^2\right] \label{eq:reddiff}
\end{equation}
where we absorb $\sigma_{\bm{\eta}}^2$ into $\lambda_t$, and $\lambda_t$ is a time-dependent weighting, for which \citet{mardani2023variational} propose a signal-to-noise ratio (SNR)-based formulation. 

\section{Theory}
\label{sec:theory}

We develop a theoretical framework that yields invariant reparameterizations for diffusion-based inverse solvers.

\subsection{Noise schedule hyperparameters}\label{subsec:noise_schedule}
\begin{figure}[h]
  \centering
  \includegraphics[width=.85\linewidth]{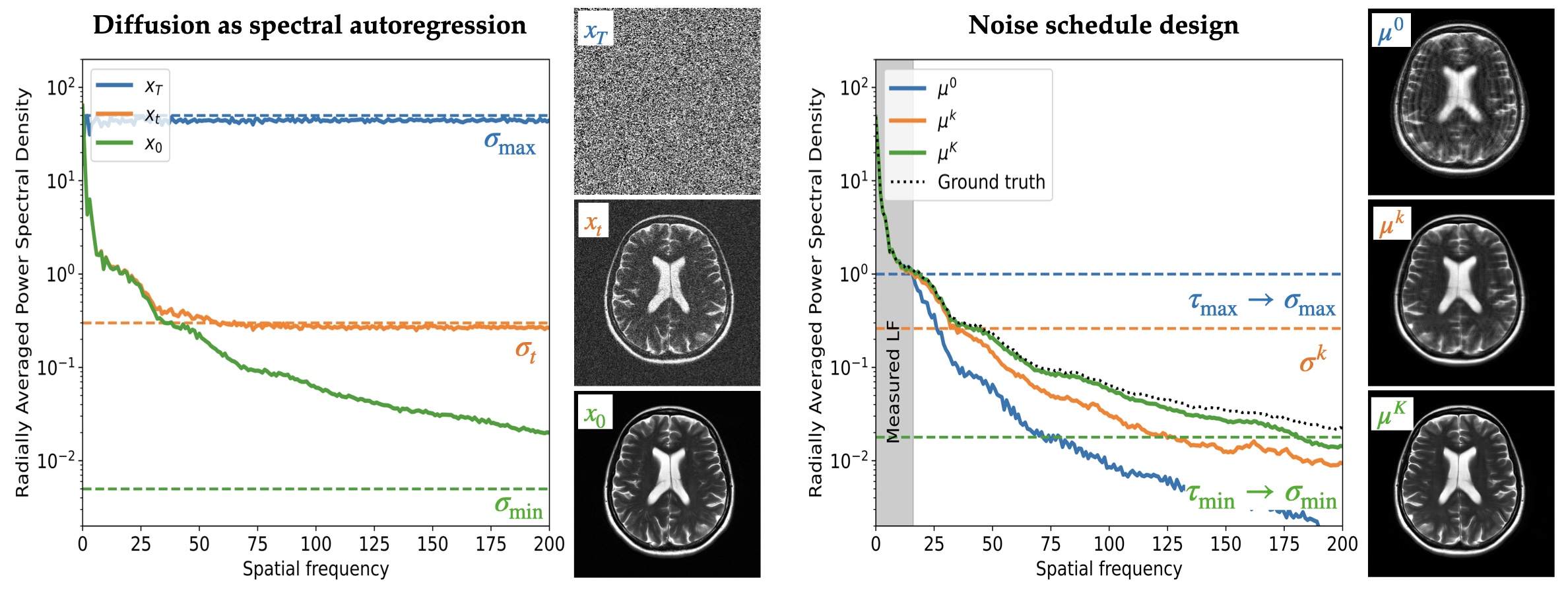}
  \caption[Noise schedule]{Noise schedule design from a spectral auto-regression perspective. \textbf{(Left)} Due to the power-law spectrum of natural images \cite{van1996modelling}, reverse diffusion recovers low-frequency components first and progressively reconstructs higher frequencies conditioned on them \cite{dieleman2024spectral, falck2025fourier}. \textbf{(Right)} This perspective motivates adapting the start and stop noise levels to measured low-frequency content and the dataset noise floor.}
  \label{fig:noise_schedule}
\end{figure}

While noise schedule design has been extensively studied for unconditional generation \cite{nichol2021improved, song2021denoising, karras2022elucidating}, its extension to inverse problems remains largely unexplored. We develop principled guidelines for selecting $\sigma_{\max}$ and $\sigma_{\min}$ by viewing reverse diffusion as spectral auto-regression in the inverse problem setting (Figure~\ref{fig:noise_schedule}). Additionally, we derive tolerance-based reparameterizations $(\tau_{\max}, \tau_{\min})$ that enable robust parameter reuse across inverse problems by analyzing the estimation error covariance . 

\paragraph{Setup.}
In Variance Exploding (VE) diffusion, we observe $\bm{x}_t= \bm{x}_0 + \sigma_t \bm{n}$, $\,\bm{n} \!\sim\!\mathcal N(0,\bm{I})$. Let $\bm{\Sigma} \succeq 0$ be the prior covariance, the LMMSE-optimal estimator yields the covariance posterior map:
\begin{equation}
\label{eq:posterior-map}
\Phi(\bm{\Sigma};\,\sigma_t) \;=\; \bm{\Sigma} - \bm{\Sigma}(\bm{\Sigma}+\sigma_t^2 \mathbf{I})^{-1}\bm{\Sigma}.
\end{equation}

\subsubsection{Start noise level $\sigma_{\max}$}
Assume the low-frequency (LF) content of $\bm{x}_0$ is known exactly, and let the prior $\bm{\Sigma}_{H\mid L}$ be the covariance of the high-frequency (HF) components conditioned on LF.  
Define the covariance residual
$
    \bm\Delta^{\max}_{t} := \bm{\Sigma}_{H\mid L} - \Phi(\bm{\Sigma}_{H\mid L};\,\sigma_t) \succeq 0,
$
and seek the region where this residual is small, meaning that the choice of $\sigma_{t}$ has negligible effect on estimation.  
We therefore start the reverse diffusion process at a noise level where the first step reduces the covariance by at most a fraction $\tau_{\max}\!\in(0,1)$ of the prior covariance.  

\begin{theorem}[Start bound]
\label{thm:sigmamax}
If $\bm\Delta^{\max}_t \preceq \tau_{\max}\,\bm{\Sigma}_{H\mid L}$, then
\begin{equation}
\label{eq:sigmamax}
\boxed{
\sigma_{\max}^2 \;\geq\; \frac{1-\tau_{\max}}{\tau_{\max}}\;\nu_{\max}\!\big(\bm{\Sigma}_{H\mid L}\big),}
\end{equation}
\end{theorem}
where $\nu_{\max}(\cdot)$ denotes the maximum eigenvalue.
\begin{corollary}[WSS/PSD]
\label{cor:sigmamax}
    Under a WSS model with ideal LF mask, ${\bm{\Sigma}}_{H\mid L}=\mathrm{diag}(S(\omega):\omega\!\in\!\Omega_H)$, where $S(\omega)$ denotes the power spectral density of the signal.  
In this case,
\begin{equation}
\label{eq:sigmamax-psd}
\sigma_{\max}^2 \;\geq\; \frac{1-\tau_{\max}}{\tau_{\max}}\;\max_{\omega\in\Omega_H} S(\omega).
\end{equation}\end{corollary}
The proofs are provided in Appendix \ref{app:thm-sigmamax} and \ref{app:cor-sigmamax}.
In practice, we choose $\sigma_{\max}$ at the lower bound, since larger values provide little reduction in error covariance (smaller $\Delta^{\max}_t$) but increase the number of reverse diffusion steps.
Expressing the bound in terms of the tolerance $\tau_{\max}$ ensures natural scaling with the spectral density and yields a more robust criterion.

\subsubsection{Stop noise level $\sigma_{\min}$}
Assume the data has an additive white noise floor $\sigma_\mathbf{s}^2$, and let $\Sigma_\mathbf{s}$ denote the covariance of the clean signal.  
The best attainable posterior (after removing VE diffusion noise) is 
$
\bm{\Sigma}_0 := \Phi(\bm{\Sigma}_\mathbf{s} ; \sigma_\mathbf{s}).
$
Define covariance residual
$
\bm{\Delta}_t^{\min} := \Phi(\bm{\Sigma}_{\mathbf{s}} ; \sqrt{\sigma^2_\mathbf{s} + \sigma^2_t}) - \bm{\Sigma}_0  \succeq 0,
$
and seek the region where this residual is small, i.e., the current posterior is sufficiently close to the best attainable posterior.  
We stop the reverse diffusion once this distance is less than a fraction $\tau_{\min} \in (0,1)$.  

\begin{theorem}[Stop bound]
\label{thm:sigmamin}
If $\bm{\Delta}_t^{\min} \preceq \tau_{\min}\,\bm{\Sigma}_{0}$ and
$\nu_{\max}(\bm{\Sigma}_{\mathbf{s}}) > \tau_{\min}\,\sigma_{\mathbf{s}}^2$, then
\begin{equation}
\label{eq:sigmamin}
\boxed{\sigma_{\min}^2 \;\leq\; \kappa(\sigma_\mathbf{s}) :=
\frac{\tau_{\min}\,\sigma_\mathbf{s}^2\,(\nu_{\max}(\bm{\Sigma}_{\mathbf{s}})+\sigma_\mathbf{s}^2)}
     {\nu_{\max}(\bm{\Sigma}_{\mathbf{s}})-\tau_{\min}\,\sigma_\mathbf{s}^2}},
\end{equation}
If $\nu_{\max} \leq \tau_{\min}\,\sigma_{\mathbf{s}}^2$, the bound is vacuous.
\end{theorem}


\begin{corollary}[High-SNR regime]
\label{cor:sigmamin}
If $\nu_{\max}\!\gg\!\sigma_\mathbf{s}^2$, then $\kappa(\sigma_\mathbf{s})\!\approx\!\tau_{\min}\,\sigma_\mathbf{s}^2$, so that 
\begin{equation}
\label{eq:sigmamin-approx}
    \sigma_{\min}^2 \lesssim \tau_{\min}\sigma_\mathbf{s}^2.
\end{equation}
\end{corollary} 

The proofs are provided in Appendix \ref{app:thm-sigmamin} and \ref{app:cor-sigmamin}.
In practice, we set $\sigma_{\min}$ at the upper bound, since smaller values already yield a posterior sufficiently close to the best attainable.  
Expressing the bound in terms of $\tau_{\min}$ ensures natural scaling with the dataset’s spectral characteristics, yielding a robust stopping criterion.

\subsection{Sampler hyperparameters}
\label{sec:first-ord-reparam}
Similarly, we propose a reparameterization of the sampler hyperparameters for the RED-diff objective in \eqref{eq:reddiff}. We show the usefulness of our proposed parameterization from an invariance and common descent direction \cite{fliege2000steepest}. Originally, RED-diff solves:
\[
\min_{\bm\mu}\; f(\bm\mu)+\mathbb{E}_t[\lambda_t\,g(\bm\mu;t)],
\quad f(\bm\mu)=\| \bm{y} - \mathcal{A} (\bm\mu)\|_2^2.
\]
At iteration $k$, let $\nabla f^k=\nabla f(\bm\mu^k)$, $\nabla g^k=\nabla g(\bm\mu^k;t^k)$, $\lambda^k = \lambda_{t_k}$. Then the stochastic gradient descent (SGD) update is
\begin{equation}
\label{eq:SGD-update}
    \bm{\mu}^{k+1} \leftarrow \bm{\mu}^k - \alpha^k \left( \nabla f^k + \lambda^k \nabla g^k \right)
\end{equation}

\paragraph{Proposed reparameterization.}
Defining 
\mbox{$r^k:=\|\nabla f^k\|_2/\|\nabla g^k\|_2$}, 
we use
\mbox{$\alpha^{k}={\hat{\alpha}^k}/{\|\nabla f^k\|_2}$}, 
\mbox{$\lambda^{k}=\hat{\lambda}\,r^k$} 
with $\hat{\alpha}^k,\hat{\lambda}>0$, yielding the \emph{unit-gradient} update
\begin{equation}
\Delta \bm\mu^k
=-\hat{\alpha}^k\!\left(
\frac{\nabla f^k}{\|\nabla f^k\|_2}
+\hat{\lambda}\,\frac{\nabla g^k}{\|\nabla g^k\|_2}
\right).
\label{eq:ugd}
\end{equation}

\noindent
\begin{minipage}[t]{0.51\linewidth}
\vspace{0pt}
\begin{theorem}[Re-scaling invariance]
\label{thm:invariance}
Let $\phi_f,\phi_g:\mathbb{R}\to\mathbb{R}$ be strictly increasing $C^1$ functions and set $\widetilde f=\phi_f\!\circ f$, $\widetilde g=\phi_g\!\circ g$. 
Applying \eqref{eq:ugd} to $(\widetilde f,\widetilde g)$ yields the same update as with $(f,g)$; the step’s direction and length are invariant to monotone reparameterizations. Proof in Appendix \ref{app:thm-invariance}.
\end{theorem}

\begin{corollary}[Practical advantage over RED-diff]
\label{cor:practical}
In RED-diff, the prior gradient often can be modeled as  $\nabla g(\bm\mu;t)=\phi'(t)\,q(\bm\mu;t)$ with a noise-level factor $\phi'(t)$ (e.g., via $\sigma_t$). 
Normalization cancels this factor:
\[
\frac{\nabla g(\bm\mu;t)}{\|\nabla g(\bm\mu;t)\|_2}=\frac{q(\bm{\mu};t)}{\|q(\bm{\mu};t)\|_2}.
\]
Thus the data–prior balance is the constant $\hat{\lambda}$ across all $t$, eliminating the need to handcraft/tune an SNR-based schedule $\lambda_t$ and avoiding searches over functional forms (linear, square-root, etc.).
\end{corollary}
\end{minipage}\hfill
\begin{minipage}[t]{0.45\linewidth}
\vspace{0pt}
  \centering
  \includegraphics[width=\linewidth]{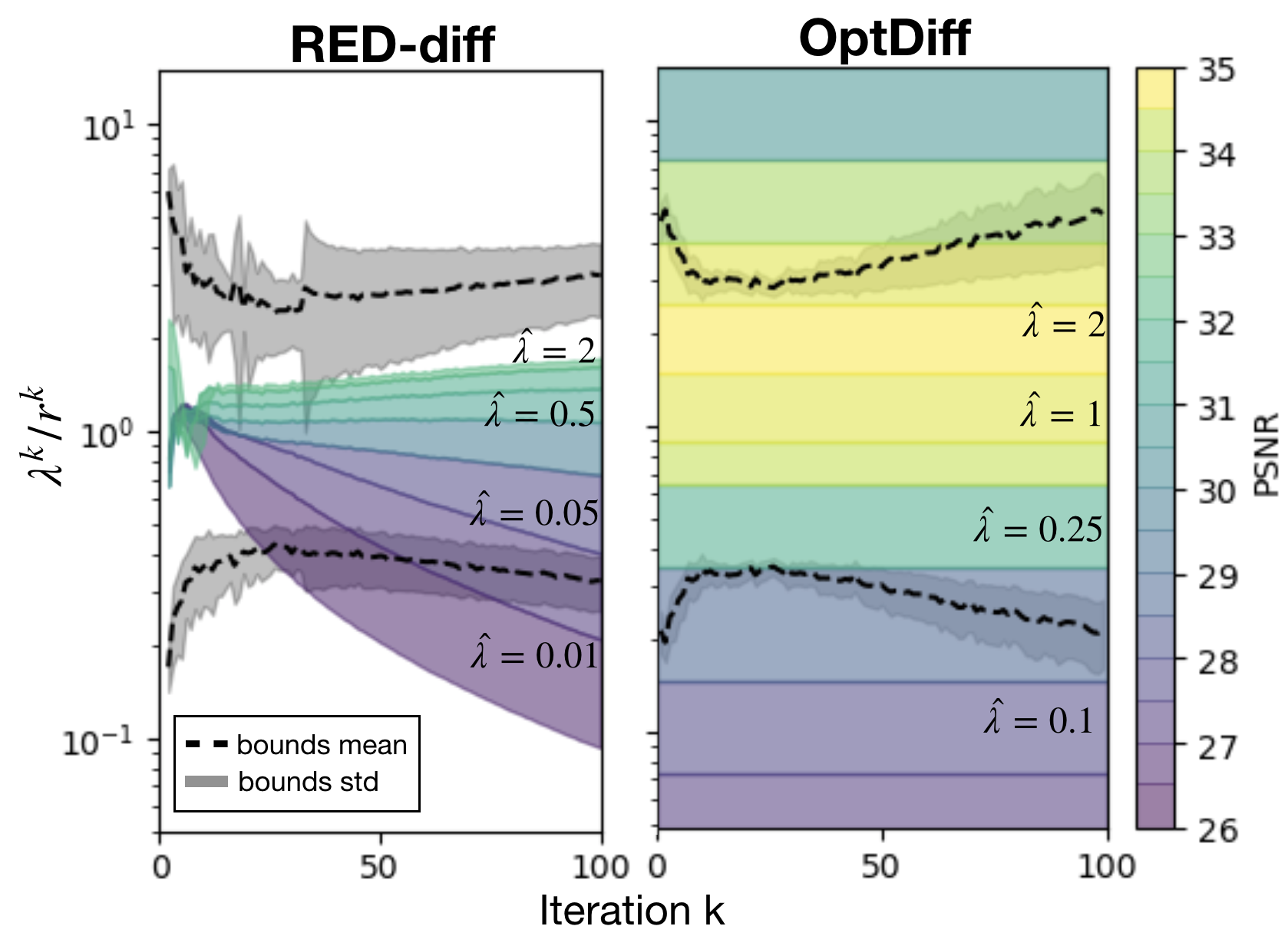}
  \captionof{figure}[Common descent]{
Common descent condition (Theorem~\ref{thm:descent}).
\textbf{(Left)} Standard $\lambda^k$ SNR-weighting violates condition bounds, resulting in degraded performance.
\textbf{(Right)} OptDiff yields consistent $\lambda^k/r^k$ trajectories, simplifying tuning and improving PSNR. 
  } \label{fig:common_descent}
\end{minipage}
\par\medskip

\begin{remark}[Interpretation]
The update depends only on the directions of $\nabla f^k$ and $\nabla g^k$, since both gradients are normalized. 
This cancels out any effect of scaling (e.g., gains, noise levels), so a single $\hat{\lambda}$ generalizes across diffusion levels.  

This idea is related to \emph{normalized SGD} methods \cite{murray2017ngd, cutkosky2020momnorm}, which rescale updates to remove dependence on gradient magnitudes, and to the \emph{Muon} optimizer \cite{jordan2024muon}, which enforces normalization and orthogonalization of updates to stabilize training.  
Our method brings these normalization principles to the diffusion-based composite objective: instead of normalizing a single gradient, we normalize \emph{both components} (data consistency and diffusion prior) so that their relative contribution is controlled solely by the trade-off parameter~$\hat{\lambda}$.
\end{remark}

\begin{theorem}[Common descent condition]
\label{thm:descent}
Let $d^k=\nabla f^k+\lambda^k\nabla g^k$ where $\lambda^k>0$ and 
$\cos\vartheta^k=\tfrac{\langle \nabla f^k,\nabla g^k\rangle}{\|\nabla f^k\|\,\|\nabla g^k\|}$.
Then common descent ($\langle \nabla f^k,d^k\rangle>0$ and $\langle \nabla g^k,d^k\rangle>0$) holds if
\[
\boxed{
\begin{cases}
\cos\vartheta^k \ge 0: & \lambda^k > 0 \quad (\text{all positive }\lambda^k\text{ are valid}),\\[6pt]
\cos\vartheta^k < 0: & \lambda^k \in \big( r^k(-\cos\vartheta^k),\; r^k(-1/\cos\vartheta^k) \big).
\end{cases}}
\]
\end{theorem}

\begin{corollary}[Simplifications under two-sided normalization]
\label{cor:normalized-rules}
Under \eqref{eq:ugd}, $\lambda^k=\hat{\lambda}\,r^k$ with global $\hat{\lambda}>0$, and Theorem~\ref{thm:descent} reduces to
\[
\boxed{ \hat{\lambda} \in \big( -\cos\vartheta^k,\; -1/\cos\vartheta^k \big) 
\quad \text{whenever } \cos\vartheta^k < 0. }
\]

The case $\cos\vartheta^k \ge 0$ is equivalent to Theorem \ref{thm:descent}. 
\end{corollary}
The proof is provided in Appendix \ref{app:thm-descent}. We display averaged common descent bounds and different values for $\hat\lambda$ in Figure \ref{fig:common_descent}. A single global $\hat{\lambda}$ can guarantee common descent and performance drops whenever $\hat\lambda$ falls out of bounds.

\section{OptDiff Pipeline}

The OptDiff pipeline combines invariant reparameterizations (Section~\ref{sec:theory}) with the RED-diff optimization perspective to enable a simplified and robust strategy for hyperparameter tuning. We demonstrate its effectiveness by designing a family of efficient optimizers for inverse problems. Below, we outline the pipeline for a given inverse problem setup (e.g., MRI reconstruction with acceleration $R=8$) and a chosen optimizer (e.g., vanilla gradient descent).

\subsection{Hyperparameter tuning}
A central component of OptDiff is a greedy-optimal hyperparameter scheme that uses a small tuning set (5 to 10 samples), similar to standard grid search, but additionally decouples noise schedule and sampler hyperparameters, and constrains the search space. This procedure also provides an upper-bound performance reference and enables the integration of convex optimization tools for accelerated posterior sampling (Section~\ref{sec:ablations}).

\paragraph{Greedy-optimal sampler hyperparameters.}
\label{subsec:opt_weights}

At each step with update rule from Eq. \ref{eq:SGD-update}, define $
    \bm{U}^k := 
    \begin{bmatrix}
        \nabla f^k & \nabla g^k 
    \end{bmatrix}
$ and $\bm{w}^k := 
    \begin{bmatrix}
        \alpha^k & \alpha^k \lambda^k 
    \end{bmatrix}^{\mathsf{T}}$, $\bm{x}^*$ is the corresponding ground-truth image from the small tuning dataset.
We estimate the greedy-optimal hyperparameters $\bm{\hat{w}}^k$  as follows:
\begin{equation}
    \bm{\hat{w}}^k = \argmin_{\bm{w} \in \mathbb{R}^2_+} \| \bm{x}^* - ( \bm{\mu}^k - \bm{U}^k \bm{w} ) \|_2^2,
\end{equation}
which can be solved via non-negative least squares (NNLS). 
This methodology is flexible and can be extended to incorporate additional optimization techniques, e.g. momentum and preconditioning. Details are provided in Appendix \ref{app:opt-weights}.

\paragraph{Noise schedule tuning.}
We perform a grid search over the tolerance parameters $\tau_{\min}$ and $\tau_{\max}$, which define the noise schedule bounds $\sigma_{\min}$ and $\sigma_{\max}$ via \eqref{eq:sigmamax-psd} and \eqref{eq:sigmamin-approx}.  
For practical implementation, we assume a WSS signal model and compute radially averaged power spectral densities. During this step, the greedy-optimal sampler hyperparameters are used to decouple sampler effects, ensuring a fair and consistent comparison across candidate noise schedules.

\paragraph{Sampler hyperparameters tuning.}
Once the noise schedule is fixed, we perform a reduced grid search over the sampler hyperparameters $\hat{\alpha}^k$ and $\hat{\lambda}$. Compared to \citet{mardani2023variational}, our invariant reparameterization eliminates the need to select a heuristic SNR-based schedule for $\lambda_t$, reducing the dimensionality of the search space. In practice, the optimal hyperparameter curves also serve as reference points, allowing the grid search to be restricted to a small neighborhood around these values. We further demonstrate the robustness of $\hat{\lambda}$ to tuning-set size and image selection through a sensitivity analysis in Appendix~\ref{app:opt-weights} (Table~\ref{tab:tuning_sensitivity}).

\subsection{Inference}
The selected hyperparameters can be directly applied to the test dataset and reused across different setups without further tuning, as shown in Figure~\ref{fig:invariance}. In particular, the same hyperparameters generalize across varying calibration regions, noise levels, accelerations, and sampling masks for the MRI reconstruction  (Figure \ref{fig:invariance} and Table \ref{tab:results}). In practice, only the step-size hyperparameter required minor adjustment between setups.

\section{Experiments}
%
%

\subsection{Invariance tests}
\label{sub:invariance}
We evaluate the invariance property of our proposed reparameterizations using the fastMRI \cite{knoll2020fastmri} dataset with the MRI reconstruction and deblurring tasks. We conducted the experiments on 10 slices from 10 different patients from the validation subset. The greedy-optimal sampler hyperparameters described in Subsection \ref{subsec:opt_weights} were used to analyze the impact of each hyperparameter in isolation.

\paragraph{Noise schedule hyperparameters.}
%
\begin{itemize}
    \item \textbf{Start noise level ($\sigma_{\max}$).}
From \eqref{eq:sigmamax}-\eqref{eq:sigmamax-psd}, the proposed $\sigma_{\max}$ depends on the low-frequency components through the conditional covariance $\bm{\Sigma}_{H|L}$ and, under the WSS assumption, on the power spectral density. Different levels of acceleration and blurring
modify  $\bm{\Sigma}_{H|L}$, thereby shifting the proposed $\sigma_{\max}$. To assess invariance, we vary reconstruction and blurring parameters, and compare the $\sigma_{\max}$ values with their reparameterized counterparts $\tau_{\max}$.
    \item  \textbf{Stop noise level $\sigma_{\min}$.}
Similarly, Eq. \ref{eq:sigmamin} shows that the proposed $\sigma_{\min}$ depends on the measurement noise floor. We test invariance by varying the measurement noise level and comparing $\sigma_{\min}$ with the reparameterized $\tau_{\min}$.
\end{itemize}

\begin{table*}[!t] 
\caption{Evaluation of noise schedule generalization across baseline solvers. Replacing the standard schedule with the optimized one improves reconstruction quality in most cases. Our method achieves the best overall performance.}
\label{tab:noise-sched-gen}
\centering
\resizebox{\textwidth}{!}{%
\begin{tabular}{c cc cc ccc ccc}
\toprule
\multirow{3}{*}{\textbf{Method}} 
& \multicolumn{4}{c}{\textbf{MRI Reconstruction R=8 (fastMRI)}} 
& \multicolumn{6}{c}{\textbf{Deblurring (ImageNet)}} \\
\cmidrule(lr){2-5} \cmidrule(lr){6-11}
& \multicolumn{2}{c}{\textbf{Standard schedule}} 
& \multicolumn{2}{c}{\textbf{Proposed schedule}} 
& \multicolumn{3}{c}{\textbf{Standard schedule}} 
& \multicolumn{3}{c}{\textbf{Proposed schedule}} \\
\cmidrule(lr){2-3} \cmidrule(lr){4-5} 
\cmidrule(lr){6-8} \cmidrule(lr){9-11}
& \textbf{PSNR} $\uparrow$ & \textbf{SSIM} $\uparrow$
& \textbf{PSNR} $\uparrow$ & \textbf{SSIM} $\uparrow$
& \textbf{PSNR} $\uparrow$ & \textbf{SSIM} $\uparrow$ & \textbf{LPIPS} $\downarrow$
& \textbf{PSNR} $\uparrow$ & \textbf{SSIM} $\uparrow$ & \textbf{LPIPS} $\downarrow$ \\
\midrule

DPS \cite{chung2022diffusion}
& $28.94 \pm 1.54$ & $0.774 \pm 0.066$
& $30.43 \pm 1.85$ & $0.803 \pm 0.046$
& $27.51 \pm 4.92$ & $0.734 \pm 0.163$ & $0.296 \pm 0.162$ 
& $38.17 \pm 3.96$ & $0.967 \pm 0.013$ & $0.007 \pm 0.005$ 
\\

DDRM \cite{kawar2022denoising}
& \cellcolor{gray!20} & \cellcolor{gray!20}
& \cellcolor{gray!20} & \cellcolor{gray!20}
& $36.24 \pm 3.23$ & $0.953 \pm 0.019$ & $0.024 \pm 0.013$ 
& $39.83 \pm 3.47$ & $0.976 \pm 0.010$ & $0.009 \pm 0.007$ 
\\

Pi-GDM \cite{song2023pseudoinverse}
& $28.88 \pm 1.27$ & $0.734 \pm 0.078$
& $29.29 \pm 1.56$ & $0.808 \pm 0.052$
& $33.85 \pm 2.84$ & $0.928 \pm 0.029$ & $0.027 \pm 0.024$ 
& $32.51 \pm 2.42$ & $0.899 \pm 0.029$ & $0.045 \pm 0.020$ 
\\

RED-diff \cite{mardani2023variational}
& $31.92 \pm 1.44$ & $0.840 \pm 0.034$
& $33.38 \pm 2.10$ & $0.888 \pm 0.042$
& $35.94 \pm 4.03$ & $0.955 \pm 0.020$ & $0.027 \pm 0.016$ 
& $40.05 \pm 3.21$ & $0.977 \pm 0.008$ & $\mathbf{0.003 \pm 0.002}$ 
\\


DAPS \cite{zhang2025improving}
& $31.66 \pm 1.49$ & $0.870 \pm 0.042$
& $31.06 \pm 2.44$ & $0.788 \pm 0.066$
& $35.03 \pm 2.27$ & $0.921 \pm 0.017$ & $0.051 \pm 0.035$ 
& $35.96 \pm 1.27$ & $0.926 \pm 0.024$ & $0.048 \pm 0.041$ 
\\

\midrule
OptDiff (Ours)
& \cellcolor{gray!20} & \cellcolor{gray!20}
& $\mathbf{34.93 \pm 1.38}$ & $\mathbf{0.921 \pm 0.029}$
& \cellcolor{gray!20} & \cellcolor{gray!20} & \cellcolor{gray!20} 
& $\mathbf{41.04 \pm 3.41}$ & $\mathbf{0.982 \pm 0.006}$ & $0.004 \pm 0.002$ 
\\

\bottomrule
\end{tabular}%
}

\label{tab:combined_results}

\par\vspace{\floatsep}
\centering

\caption{
Ablation studies to evaluate performance of (a) our invariant reparameterizations and (b) integration of advanced optimization tools. Our $\lambda$-reparameterization outperforms all SNR-based baselines. Momentum and preconditioning improve reconstruction quality.
}
\label{tab:ablation}

\renewcommand{\arraystretch}{0.95}
\setlength{\tabcolsep}{3pt}
\scriptsize
\resizebox{\linewidth}{!}{%
\begin{tabular}{@{}cc@{}}
\begin{tabular}{c}
\multicolumn{1}{c}{\textbf{(a) Invariant reparameterization ablations}}\\
\begin{tabular}{ccccc}
\toprule
& \multicolumn{2}{c}{\textbf{Standard schedule}} &
\multicolumn{2}{c}{\textbf{Proposed schedule}} \\
\cmidrule(lr){2-3} \cmidrule(lr){4-5}

\textbf{$\lambda$ method}
& \textbf{PSNR} $\uparrow$ & \textbf{SSIM} $\uparrow$
& \textbf{PSNR} $\uparrow$ & \textbf{SSIM} $\uparrow$ \\
\midrule

Greedy-optimal
& $33.13 \pm 1.22$
& $0.904 \pm 0.020$
& $34.78 \pm 1.24$
& $0.941 \pm 0.021$ \\
\midrule

Constant
& $30.53 \pm 1.31$
& $0.732 \pm 0.040$
& $33.56 \pm 1.49$
& $0.910 \pm 0.030$ \\

Linear
& $30.40 \pm 1.35$
& $0.846 \pm 0.041$
& $34.39 \pm 1.24$
& $0.928 \pm 0.021$ \\

Square
& $29.62 \pm 1.18$
& $0.834 \pm 0.024$
& $32.35 \pm 1.29$
& $0.910 \pm 0.025$ \\

Sqrt.
& $32.19 \pm 1.24$
& $0.835 \pm 0.034$
& $34.13 \pm 1.31$
& $0.921 \pm 0.025$ \\

Log
& $31.98 \pm 1.82$
& $0.893 \pm 0.025$
& $34.36 \pm 1.25$
& $0.928 \pm 0.021$ \\

\midrule

Ours
& $\mathbf{32.87 \pm 1.31}$
& $\mathbf{0.895 \pm 0.026}$
& $\mathbf{34.51 \pm 1.37}$
& $\mathbf{0.927 \pm 0.025}$ \\

\bottomrule
\end{tabular}
\end{tabular}
&
\hspace{7pt}
\begin{tabular}{c}
\multicolumn{1}{c}{\textbf{(b) Optimizer ablations}}\\
\begin{tabular}{llcccc}
\toprule

\multicolumn{2}{c}{} &
\multicolumn{2}{c}{\textbf{NFE 20}} &
\multicolumn{2}{c}{\textbf{NFE 100}} \\
\cmidrule(lr){3-4} \cmidrule(lr){5-6}

\multicolumn{2}{c}{} &
\textbf{PSNR} $\uparrow$ & \textbf{SSIM} $\uparrow$ &
\textbf{PSNR} $\uparrow$ & \textbf{SSIM} $\uparrow$ \\
\midrule

\multicolumn{2}{c}{RED-diff}
& $29.7 \pm 1.2$
& $0.84 \pm 0.03$
& $32.2 \pm 1.2$
& $0.84 \pm 0.03$ \\

\midrule

\multirow{3}{*}{\shortstack[l]{Greedy-Optimal}}
& Vanilla
& $30.1 \pm 1.7$
& $0.89 \pm 0.04$
& $33.8 \pm 1.4$
& $0.92 \pm 0.02$ \\

& Moment
& $31.8 \pm 2.1$
& $0.89 \pm 0.04$
& $34.6 \pm 1.4$
& $0.93 \pm 0.02$ \\

& Precond
& $\mathbf{32.3 \pm 2.2}$
& $\mathbf{0.90 \pm 0.05}$
& $\mathbf{34.6 \pm 1.4}$
& $\mathbf{0.93 \pm 0.02}$ \\

\midrule

\multirow{3}{*}{\shortstack[l]{OptDiff}}
& Vanilla
& $29.9 \pm 1.7$
& $0.88 \pm 0.04$
& $33.4 \pm 1.6$
& $0.91 \pm 0.03$ \\

& Moment
& $31.7 \pm 2.1$
& $0.87 \pm 0.05$
& $34.4 \pm 1.3$
& $0.92 \pm 0.02$ \\

& Precond
& $\mathbf{32.3 \pm 2.1}$
& $\mathbf{0.89 \pm 0.05}$
& $\mathbf{34.5 \pm 1.4}$
& $\mathbf{0.92 \pm 0.02}$ \\

\bottomrule
\end{tabular}
\end{tabular}
\end{tabular}}

\vspace{-5pt}
\end{table*}

\begin{table}[!t]
\centering
\caption{
Comparison with baselines using NFE-100. Our method consistently achieves the best or comparable performance across all task--dataset pairs. Best results are shown in bold, and second-best results are underlined.}
\label{tab:results}
\resizebox{\linewidth}{!}{%
\begin{tabular}{c l ccc ccc}

\toprule
& & \multicolumn{6}{c}{\textbf{fastMRI}} \\
& & \multicolumn{3}{c}{\textbf{Acceleration R=4}} & \multicolumn{3}{c}{\textbf{Acceleration R=8}} \\
\cmidrule(lr){3-5} \cmidrule(lr){6-8}
\textbf{Task} & \textbf{Method} 
& \textbf{PSNR} $\uparrow$ & \textbf{SSIM} $\uparrow$ & \cellcolor{gray!20}{}
& \textbf{PSNR} $\uparrow$ & \textbf{SSIM} $\uparrow$ & \cellcolor{gray!20}{} \\
\midrule

\multirow{7}{*}{\shortstack{MRI recon}}

& DPS \cite{chung2022diffusion}
& $35.82 \pm 1.46$ & $0.775 \pm 0.069$ & \cellcolor{gray!20}{}
& $28.94 \pm 1.54$ & $0.774 \pm 0.066$ & \cellcolor{gray!20}{} \\

& DiffPIR \cite{zhu2023denoising}
& $30.33 \pm 0.97$&  ${0.691 \pm 0.077}$& \cellcolor{gray!20}{}
& $27.17 \pm 1.08$ & $0.650 \pm 0.072$ & \cellcolor{gray!20}{} \\

& Pi-GDM \cite{song2023pseudoinverse}
& $30.69 \pm 1.37$ & $0.753 \pm 0.086$ & \cellcolor{gray!20}{}
& $28.88 \pm 1.27$ & $0.734 \pm 0.078$ & \cellcolor{gray!20}{} \\

& RED-diff \cite{mardani2023variational}
& $\underline{36.79 \pm 1.43}$ & $0.872 \pm 0.035$ & \cellcolor{gray!20}{}
& $\underline{31.92 \pm 1.44}$ & $0.840 \pm 0.034$ & \cellcolor{gray!20}{} \\

& DDS \cite{chung2024decomposed}
& $33.72 \pm 1.14$ & $\underline{0.930 \pm 0.030}$ & \cellcolor{gray!20}{}
& $28.92 \pm 1.28$ & $0.828 \pm 0.027$ & \cellcolor{gray!20}{} \\

& DAPS \cite{zhang2025improving}
& $36.50 \pm 1.63$ & $0.915 \pm 0.033$ & \cellcolor{gray!20}{}
& $31.66 \pm 1.49$ & $\underline{0.870 \pm 0.042}$ & \cellcolor{gray!20}{} \\

\cmidrule(lr){2-8}

& OptDiff (Ours)
& $\mathbf{38.45 \pm 1.58}$ & $\mathbf{0.948 \pm 0.023}$ & \cellcolor{gray!20}{}
& $\mathbf{34.93 \pm 1.38}$ & $\mathbf{0.921 \pm 0.029}$ & \cellcolor{gray!20}{} \\
\bottomrule

\toprule
& & \multicolumn{3}{c}{\textbf{ImageNet}} & \multicolumn{3}{c}{\textbf{FFHQ}} \\
\cmidrule(lr){3-5} \cmidrule(lr){6-8}
\textbf{Task} & \textbf{Method} 
& \textbf{PSNR} $\uparrow$ & \textbf{SSIM} $\uparrow$ & \textbf{LPIPS} $\downarrow$
& \textbf{PSNR} $\uparrow$ & \textbf{SSIM} $\uparrow$ & \textbf{LPIPS} $\downarrow$ \\
\cmidrule(lr){1-8}

\multirow{7}{*}{\shortstack{Deblurring \\ kernel $6$}}

& DPS \cite{chung2022diffusion}
& $24.25 \pm 3.97$ & $0.602 \pm 0.192$ & $0.420 \pm 0.187$
& $33.00 \pm 2.50$ & $0.898 \pm 0.034$ & $0.078 \pm 0.028$ \\

& DDRM \cite{kawar2022denoising}    
& $\underline{32.71 \pm 3.82}$ & $\underline{0.908 \pm 0.045}$ & $\underline{0.039 \pm 0.017}$
& $\underline{36.96 \pm 1.76}$ & $\mathbf{0.949 \pm 0.013}$ & $\underline{0.030 \pm 0.009}$ \\

& DDNM \cite{wang2022zero}          
& $32.60 \pm 3.89$ & $0.906 \pm 0.045$ & $\mathbf{0.030 \pm 0.013}$
& $36.95 \pm 1.78$ & $\mathbf{0.949 \pm 0.013}$ & $\mathbf{0.027 \pm 0.008}$ \\

& DiffPIR \cite{zhu2023denoising}   
& $32.22 \pm 3.68$ & $0.892 \pm 0.049$ & $0.061 \pm 0.032$
& $36.58 \pm 1.76$ & $\underline{0.942 \pm 0.014}$ & $0.040 \pm 0.012$ \\

& RED-diff \cite{mardani2023variational} 
& $29.35 \pm 3.92$ & $0.821 \pm 0.092$ & $0.175 \pm 0.089$
& $35.07 \pm 1.77$ & $0.915 \pm 0.015$ & $0.063 \pm 0.020$ \\

& DAPS \cite{zhang2025improving}    
& $31.39 \pm 3.13$ & $0.853 \pm 0.047$ & $0.107 \pm 0.033$
& $34.13 \pm 1.39$ & $0.888 \pm 0.013$ & $0.088 \pm 0.021$ \\

\cmidrule(lr){2-8}

& OptDiff (Ours)     
& $\mathbf{33.28 \pm 3.74}$ & $\mathbf{0.914 \pm 0.038}$ & $0.049 \pm 0.024$
& $\mathbf{37.10 \pm 1.79}$ & $\mathbf{0.949 \pm 0.013}$ & $0.058 \pm 0.017$ \\

\midrule

\multirow{7}{*}{\shortstack{Severe deblur \\ kernel $12$}}

& DPS \cite{chung2022diffusion}
& $21.30 \pm 3.27$ & $0.484 \pm 0.200$ & $0.371 \pm 0.159$
& $25.88 \pm 6.47$ & $0.749 \pm 0.129$ & $0.159 \pm 0.115$ \\

& DDRM \cite{kawar2022denoising}    
& $27.87 \pm 3.89$ & $0.785 \pm 0.110$ & $0.156 \pm 0.074$
& $32.18 \pm 2.18$ & $0.889 \pm 0.031$ & $\underline{0.069 \pm 0.021}$ \\

& DDNM \cite{wang2022zero}
& $\underline{28.07 \pm 4.12}$ & $\mathbf{0.789 \pm 0.110}$ & $\mathbf{0.134 \pm 0.059}$
& $\underline{32.54 \pm 2.15}$ & $\mathbf{0.893 \pm 0.030}$ & $0.074 \pm 0.022$ \\

& DiffPIR \cite{zhu2023denoising}
& $27.41 \pm 3.72$ & $0.746 \pm 0.114$ & $\underline{0.145 \pm 0.064}$
& $32.05 \pm 1.95$ & $0.870 \pm 0.030$ & $\mathbf{0.060 \pm 0.019}$ \\

& RED-diff  \cite{mardani2023variational}
& $24.19 \pm 3.49$ & $0.604 \pm 0.171$ & $0.428 \pm 0.156$
& $29.42 \pm 2.23$ & $0.800 \pm 0.048$ & $0.183 \pm 0.054$ \\

& DAPS  \cite{zhang2025improving}    
& $26.28 \pm 2.80$ & $0.671 \pm 0.086$ & $0.247 \pm 0.075$
& $28.38 \pm 2.19$ & $0.786 \pm 0.055$ & $0.199 \pm 0.040$ \\

\cmidrule(lr){2-8}

& OptDiff (Ours) 
& $\mathbf{28.27 \pm 3.90}$ & $\underline{0.788 \pm 0.100}$ & $0.190 \pm 0.083$
& $\mathbf{32.61 \pm 2.22}$ & $\underline{0.891 \pm 0.031}$ & $0.127 \pm 0.034$ \\

\midrule

\multirow{7}{*}{Super-res $4\times$}

& DPS \cite{chung2022diffusion}
& $23.72 \pm 4.90$ & $0.618 \pm 0.209$ & $0.299 \pm 0.207$
& $30.50 \pm 2.62$ & $0.866 \pm 0.044$ & $\mathbf{0.071 \pm 0.025}$ \\

& DDRM \cite{kawar2022denoising}     
& $25.84 \pm 3.86$ & $0.702 \pm 0.152$ & $0.331 \pm 0.145$
& $\underline{31.68 \pm 2.50}$ & $\underline{0.885 \pm 0.038}$ & $0.125 \pm 0.038$ \\

& DDNM \cite{wang2022zero}           
& $25.79 \pm 3.93$ & $0.703 \pm 0.152$ & $0.286 \pm 0.136$
& $30.94 \pm 2.45$ & $0.872 \pm 0.041$ & $\underline{0.085 \pm 0.028}$ \\

& DiffPIR \cite{zhu2023denoising}    
& $25.28 \pm 3.78$ & $0.682 \pm 0.152$ & $\underline{0.242 \pm 0.117}$
& $31.06 \pm 2.44$ & $0.873 \pm 0.040$ & $0.092 \pm 0.031$ \\

& RED-diff  \cite{mardani2023variational} 
& $25.75 \pm 3.82$ & $0.700 \pm 0.141$ & $0.311 \pm 0.144$
& $31.25 \pm 2.37$ & $0.875 \pm 0.037$ & $0.108 \pm 0.034$ \\

& DAPS \cite{zhang2025improving}    
& $\mathbf{26.50 \pm 3.43}$ & $\underline{0.706 \pm 0.132}$ & $\mathbf{0.239 \pm 0.067}$
& $29.63 \pm 1.73$ & $0.763 \pm 0.028$ & $0.179 \pm 0.028$ \\

\cmidrule(lr){2-8}

& OptDiff (Ours)     
& $\underline{26.07 \pm 3.96}$ & $\mathbf{0.715 \pm 0.148}$ & $0.350 \pm 0.156$
& $\mathbf{31.99 \pm 2.57}$ & $\mathbf{0.891 \pm 0.037}$ & $0.147 \pm 0.043$ \\

\midrule

\multirow{7}{*}{Super-res $16\times$}

& DPS \cite{chung2022diffusion}
& $17.95 \pm 3.02$ & $0.365 \pm 0.192$ & $\mathbf{0.388 \pm 0.125}$
& $22.62 \pm 2.28$ & $0.664 \pm 0.086$ & $0.286 \pm 0.078$ \\

& DDRM \cite{kawar2022denoising}     
& $20.35 \pm 2.77$ & $\underline{0.462 \pm 0.185}$ & $0.683 \pm 0.199$
& $\underline{23.50 \pm 2.24}$ & $\underline{0.687 \pm 0.083}$ & $0.316 \pm 0.090$ \\

& DDNM \cite{wang2022zero}
& $20.15 \pm 2.68$ & $0.460 \pm 0.183$ & $0.633 \pm 0.207$
& $22.86 \pm 2.12$ & $0.663 \pm 0.083$ & $\underline{0.249 \pm 0.079}$ \\

& DiffPIR \cite{zhu2023denoising}
& $19.66 \pm 2.61$ & $0.439 \pm 0.178$ & $0.577 \pm 0.194$
& $22.77 \pm 2.14$ & $0.658 \pm 0.085$ & $\mathbf{0.246 \pm 0.077}$ \\

& RED-diff  \cite{mardani2023variational}
& $20.34 \pm 2.76$ & $0.460 \pm 0.182$ & $0.704 \pm 0.193$
& $23.01 \pm 2.07$ & $0.663 \pm 0.082$ & $0.282 \pm 0.082$ \\

& DAPS  \cite{zhang2025improving}     
& $\mathbf{20.53 \pm 2.49}$ & $0.418 \pm 0.136$ & $\underline{0.522 \pm 0.081}$
& $22.63 \pm 1.96$ & $0.572 \pm 0.064$ & $0.358 \pm 0.046$ \\

\cmidrule(lr){2-8}

& OptDiff (Ours) 
& $\underline{20.44 \pm 2.75}$ & $\mathbf{0.468 \pm 0.184}$ & $0.704 \pm 0.186$
& $\mathbf{23.67 \pm 2.25}$ & $\mathbf{0.689 \pm 0.083}$ & $0.343 \pm 0.094$ \\
\bottomrule
\end{tabular}
}

\par\vspace{\floatsep}

  \centering
  \includegraphics[width=\linewidth]{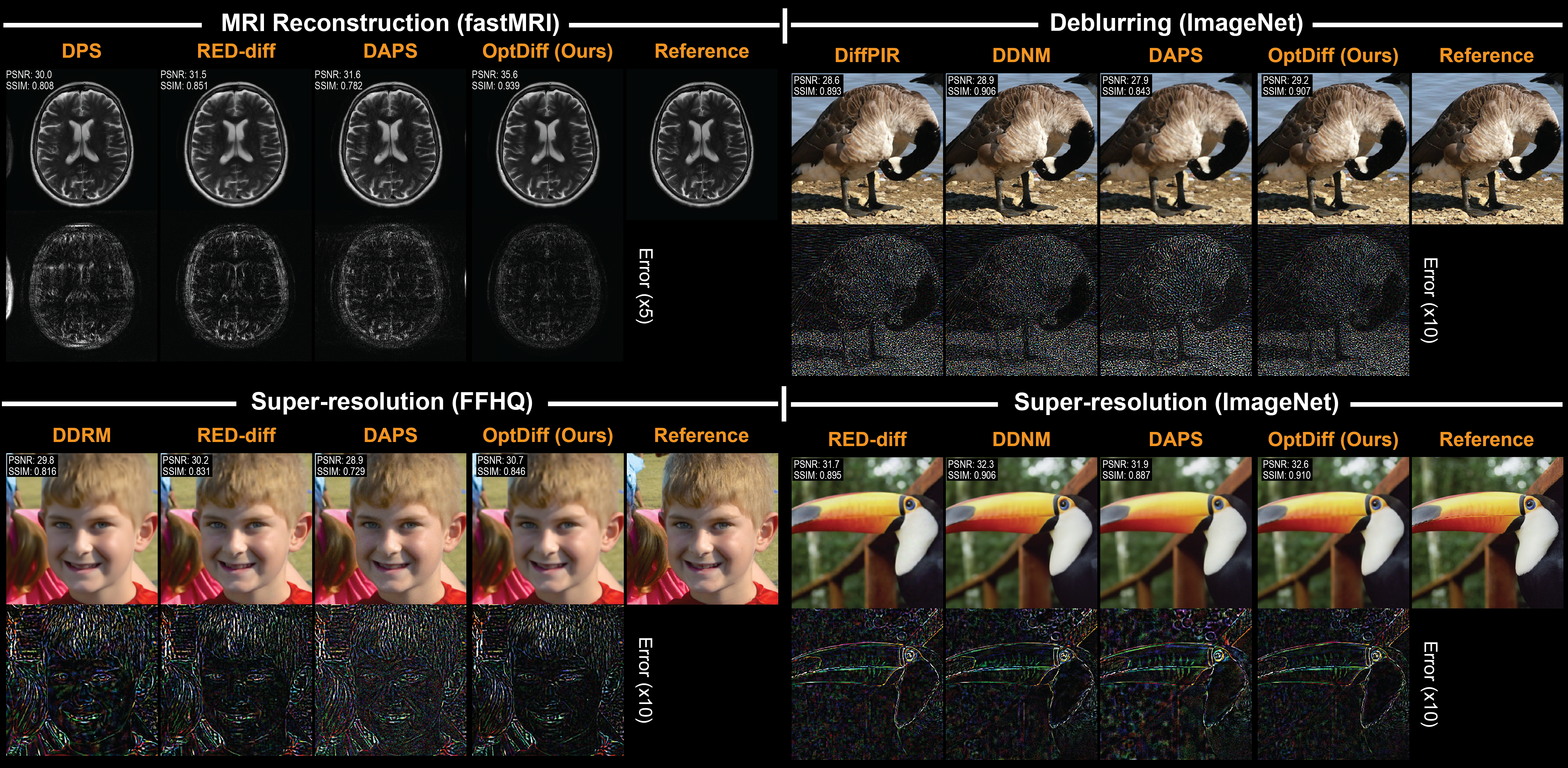}
  \captionof{figure}[Comparisons with baseline methods]{Performance comparison versus baseline methods.} \label{fig:visual_comparisons}
\end{table}

\paragraph{Noise schedule transfer across solvers}
The bounds on $\sigma_{\max}$ and $\sigma_{\min}$ in Section \ref{subsec:noise_schedule} are deliberately derived from the diffusion denoising model, independently of any particular inverse solver. This motivates testing the resulting noise schedule beyond OptDiff. We evaluate several diffusion-based inverse solvers using both their standard schedules and our proposed schedule.

\paragraph{Sampler hyperparameters.}
As discussed in Section~\ref{sec:first-ord-reparam}, inverse solving performance depends on the choice of the weighting function $\lambda^k$, which compensates for variations in the norm of $\nabla g$. Our reparameterization cancels any positive rescaling of the prior gradient, thereby removing the need for a handcrafted schedule. To test this invariance, we examine whether multiple base values $\hat{\lambda}\in\mathbb{R}$ are required across the reverse diffusion process. We sweep $\hat{\lambda}$ for both the standard update in Eq.~\ref{eq:reddiff} with a linear schedule and the normalized update in Eq.~\ref{eq:ugd}, evaluating performance over three intervals: $k \in [0,33]$, $[33,66]$, and $[66,99]$. We use the optimal step size $\alpha^k$ as described in subsection \ref{subsec:opt_weights}.

\paragraph{Results.} As shown in Figure \ref{fig:invariance}, the $\tau$-parameterizations produce well-aligned curves across conditions, whereas the original $\sigma$-parameterizations show large variability. This confirms that our reparameterization scales naturally with the spectral properties of the data and the problem setup. Moreover, Table \ref{tab:noise-sched-gen} demonstrates that the proposed noise schedule generalizes across solvers. In most cases, replacing the standard schedule with the optimized schedule improves reconstruction quality across baseline methods, while OptDiff achieves the best overall performance under the optimized schedule.
Similarly, Figure~\ref{fig:invariance} shows that with the standard formulation, the $\hat{\lambda}$ curves are dispersed across intervals, implying that multiple base values are needed. In contrast, our reparameterization aligns the curves, demonstrating that a single $\hat{\lambda}$ suffices across the entire reverse process by accounting for the dynamics of $\|\nabla g\|_2$.

\subsection{Ablations}
\label{sec:ablations}

We conduct ablations to evaluate (a) the performance of our invariant reparameterizations, i.e. $\lambda$-weighting and noise schedule design , and (b) the effect of incorporating advanced optimization tools to the RED-diff framework. As in Section~\ref{sub:invariance}, we use MRI reconstruction at acceleration factor $R=8$ with 100 iterations as the base task. For (a), we compare fixed SNR-based weightings $\lambda_k = h(\mathrm{SNR}_k)$ against our reparameterization under both the default EDM noise schedule ($\sigma_{\min}=0.002$, $\sigma_{\max}=20$) and the proposed schedule. For (b), we assess the impact of momentum and polynomial preconditioning at 20 and 100 NFE. In standard diffusion solvers, such methods typically introduce additional hyperparameters; however, our invariant formulation and hyperparameter tuning strategy enables their integration without increasing tuning complexity. 

\paragraph{Results.} Table~\ref{tab:ablation}a shows that our $\lambda$-reparameterization consistently outperforms all SNR-based baselines under both standard and optimized noise schedules, with the oracle $\lambda$ (computed using the optimal hyperparameters in Subsection \ref{subsec:opt_weights}) serving as an upper bound. Table~\ref{tab:ablation}b further demonstrates that momentum and preconditioning consistently improve reconstruction quality over the vanilla optimizer, with larger gains at 20 steps and diminishing but still positive gains at NFE-100.

\subsection{Tests with Baseline Methods}
\label{sub:comparisons}

We conduct two main experiments. First, we evaluate whether the proposed noise-schedule design generalizes to state-of-the-art diffusion-based inverse solvers beyond RED-diff and OptDiff. Second, we compare the reconstruction performance of our designed optimizers against these baseline solvers. Experiments are performed on three datasets: fastMRI, ImageNet, and FFHQ, across four inverse problems: MRI reconstruction, deblurring, super-resolution, and inpainting. Details in Appendix \ref{app:impl-details}.

\paragraph{Results.}
Figure~\ref{fig:visual_comparisons} and Table~\ref{tab:results} compare OptDiff against existing diffusion-based inverse solvers. Across tasks and datasets, OptDiff achieves the best or comparable PSNR, SSIM, and LPIPS values relative to the baselines. In addition, OptDiff provides a favorable speed--quality trade-off. Our method either matches the fastest runtime or remains competitive while achieving higher reconstruction quality.

\section{Limitations}
\label{sec:limit}

We briefly summarize the main limitations here and provide a more detailed discussion with future directions in Appendix~\ref{app:limitations}. First, the proposed noise-schedule design is most effective when the forward model preserves sufficient low-frequency information, enabling a reduced $\sigma_{\max}$. This holds for the main tasks considered in this work, but may be less effective for tasks such as inpainting.
Second, OptDiff inherits limitations from RED-diff and other optimization-based posterior samplers that produce a single reconstruction, potentially favoring pixelwise fidelity over perceptual quality in highly ill-posed tasks with multiple plausible solutions \cite{daras2024survey,blau2018perception}. This tendency may be further reinforced by our MSE-based hyperparameter tuning.
Finally, while the noise-schedule hyperparameters transfer naturally to other inverse solvers, the sampler-hyperparameters are more specific to optimization-based formulations. Extending this component to broader classes of posterior samplers is left for future work.

\section{Conclusion}
We introduced OptDiff, a framework that improves the robustness and efficiency of diffusion-based inverse solvers through principled reparameterizations and optimization tools. By inducing invariances and reducing the effective hyperparameter search space, OptDiff enables stable performance across inverse problem with minimal retuning. The results also suggest future directions for incorporating more advanced optimization techniques into diffusion-based posterior sampling \cite{nocedal2006numerical, boyd2004convex}.

\section{Acknowledgements}
This work was supported in part by the National Science
Foundation (NSF) CAREER Award under Grant CCF-2236829,
in part by the NSF under Grant CNS-2551282, in part by the National Institutes of Health under Grants R01EB036127 and 1R01AG089509-01A1,
in part by the Office of Naval Research under Grant
N00014-24-1-2164, and in part by the Defense Advanced
Research Projects Agency under Grant HR00112490441.




\bibliographystyle{unsrtnat}
\bibliography{new_bib}
\newpage
\appendix

\section{Notation glossary}
\label{app:notation}
This glossary summarizes the principal notation used in the main text and appendices. The diffusion time $t$ and the discrete iteration index $k$ are distinguished throughout; symbols introduced for individual proofs retain their local definitions.

\begingroup
\small
\renewcommand{\arraystretch}{1.15}
\begin{longtable}{@{}p{0.25\textwidth}p{0.72\textwidth}@{}}
\toprule
\textbf{Notation} & \textbf{Meaning} \\
\midrule
\endfirsthead
\toprule
\textbf{Notation} & \textbf{Meaning} \\
\midrule
\endhead
\midrule
\multicolumn{2}{r}{\footnotesize Continued on next page} \\
\endfoot
\bottomrule
\endlastfoot
\multicolumn{2}{@{}l}{\textbf{Inverse problems and diffusion}} \\
$\bm{x}_0,\bm{x}_t$ & Data image and its perturbed state at diffusion time $t$. \\
$\bm{y},\mathcal{A}$ & Measurements and measurement operator, with $\bm{y}=\mathcal{A}(\bm{x}_0)+\bm{\eta}$. \\
$n,m$ & Image and measurement dimensions, respectively. \\
$\bm{\eta},\sigma_{\bm{\eta}}^2$ & Gaussian measurement noise and its variance per component, with covariance $\sigma_{\bm{\eta}}^2\mathbf{I}$. \\
$t,T$ & Continuous diffusion time and terminal time. \\
$p_{\mathrm{data}},p_t$ & Data distribution and marginal distribution at time $t$. \\
$\nabla_{\bm{x}_t}\log p_t$ & Score of the perturbed data distribution. \\
$\bm{b}(\bm{x}_t,t),a(t)$ & Vector drift and squared diffusion amplitude in the forward SDE; the noise amplitude is $\sqrt{a(t)}$. \\
$\bm{W}_t,\bar{\bm{W}}_t$ & Forward and reverse-time Wiener processes. \\
$\bm{\epsilon}_{\bm{\theta}}(\bm{x}_t;t)$ & Noise-prediction network with learned parameters $\bm{\theta}$. \\
$\bm{n},\bm{\epsilon}$ & Standard Gaussian noise vectors in the perturbation and sampling formulations. \\
$\sigma_t$ & Standard deviation of the diffusion perturbation. \\
$\sigma_{\max},\sigma_{\min}$ & Starting and stopping noise levels of reverse diffusion. \\
\midrule
\multicolumn{2}{@{}l}{\textbf{Spectral quantities and noise-schedule criteria}} \\
$\omega,S(\omega)$ & Spatial frequency and signal power spectral density. \\
$\Omega_H$ & Set of unknown high frequencies. \\
$\bm{\Sigma},\bm{\Sigma}_{H\mid L}$ & Prior covariance and LMMSE error covariance of HF components after using exactly known LF components. \\
$\Phi(\bm{\Sigma};\sigma)$ & LMMSE error covariance for a signal with covariance $\bm{\Sigma}$ observed with additive white noise of covariance $\sigma^2\mathbf{I}$, uncorrelated with the signal. \\
$\bm{\Delta}_t^{\max}$ & Reduction from prior HF uncertainty: $\bm{\Sigma}_{H\mid L}-\Phi(\bm{\Sigma}_{H\mid L};\sigma_t)$. \\
$\tau_{\max},\tau_{\min}$ & Relative covariance tolerances, referenced to $\bm{\Sigma}_{H\mid L}$ at the start and $\bm{\Sigma}_0$ at the stop. \\
$\bm{\Sigma}_{\mathbf{s}},\sigma_{\mathbf{s}}^2$ & Clean-signal covariance and effective white noise floor variance in the image domain used in the stopping analysis. \\
$\bm{\Sigma}_0$ & In the stopping analysis, limiting LMMSE error covariance as $\sigma_t\to0$: $\Phi(\bm{\Sigma}_{\mathbf{s}};\sigma_{\mathbf{s}})$. In Appendix~\ref{app:cor-sigmamax}, this symbol instead denotes $\mathrm{Cov}(\bm{x}_0)$. \\
$\bm{\Delta}_t^{\min}$ & Excess LMMSE error covariance above $\bm{\Sigma}_0$ when the total noise variance is $\sigma_{\mathbf{s}}^2+\sigma_t^2$. \\
$\nu_{\max}(\cdot)$ & Largest eigenvalue of a covariance matrix. \\
$\kappa(\sigma_{\mathbf{s}})$ & Upper bound on $\sigma_{\min}^2$ when $\nu_{\max}(\bm{\Sigma}_{\mathbf{s}})>\tau_{\min}\sigma_{\mathbf{s}}^2$; otherwise the stopping bound is vacuous. \\
\midrule
\multicolumn{2}{@{}l}{\textbf{Sampler reparameterization and common descent}} \\
$k,\bm{\mu}^k$ & Iteration index and current reconstruction estimate, represented by the variational mean. \\
$f,g$ & Data-consistency objective and diffusion-prior regularizer. \\
$\nabla f^k,\nabla g^k$ & Gradients evaluated at the current iterate and, for $g$, the current diffusion noise level. \\
$\alpha^k,\lambda^k$ & Conventional step size and regularization weight. \\
$r^k$ & Gradient-norm ratio, $\|\nabla f^k\|_2/\|\nabla g^k\|_2$. \\
$\hat{\alpha}^k,\hat{\lambda}$ & Reparameterized step scale and trade-off weight: $\alpha^k=\hat{\alpha}^k/\|\nabla f^k\|_2$ and $\lambda^k=\hat{\lambda}r^k$. \\
$\Delta\bm{\mu}^k$ & Reconstruction update, $\bm{\mu}^{k+1}-\bm{\mu}^k$. \\
$\vartheta^k$ & Angle between the data-consistency and diffusion-prior gradients. \\
$d^k$ & Combined gradient, $\nabla f^k+\lambda^k\nabla g^k$, in the common-descent analysis; the vanilla SGD update is $-\alpha^k d^k$. \\
\midrule
\multicolumn{2}{@{}l}{\textbf{Greedy-optimal tuning and solver acceleration}} \\
$\bm{x}^*$ & Ground-truth reference image used for tuning. \\
$\bm{U}^k$ & Matrix of update directions, including momentum or preconditioned gradients where applicable. \\
$\bm{w}^k,\hat{\bm{w}}^k$ & Update coefficients and their greedy-optimal NNLS estimates. For vanilla updates, $\bm{w}^k=[\alpha^k,\alpha^k\lambda^k]^{\mathsf T}$. \\
$\bm{v}^k,\beta^k$ & Momentum state and its coefficient; $\bm{v}^{k+1}$ is subtracted from $\bm{\mu}^k$, so $\Delta\bm{\mu}^k=-\bm{v}^{k+1}$. \\
$\mathbf{A}$ & Matrix representation of a linear measurement operator. \\
$\mathrm{pol}(\mathbf{A}^{\mathsf T}\mathbf{A})$ & Polynomial preconditioner applied to the data-consistency gradient. \\
$N$, NFE & Number of diffusion steps and number of neural function evaluations. \\
\midrule
\multicolumn{2}{@{}l}{\textbf{LMMSE analysis and matrix notation}} \\
$\bm{u},\bm{v},\mathbf{B}$ & Signal, observation, and linear observation matrix in the appendix LMMSE model. \\
$\mathbf{D}$ & LMMSE linear estimator mapping observations to signal estimates. \\
$\bm{u}_L,\bm{u}_H$ & Low- and high-frequency components of a signal. \\
$\bm{\Sigma}_{\bm{u}},\bm{\Sigma}_{\bm{u}\bm{v}}$ & Covariance of $\bm{u}$ and cross-covariance of $\bm{u}$ and $\bm{v}$. \\
$\bm{\Sigma}_{\bm{u}\mid\bm{v}}$ & LMMSE error covariance for estimating $\bm{u}$ from $\bm{v}$. \\
$\mathbf{I},(\cdot)^{\mathsf T},(\cdot)^\dagger$ & Identity matrix, transpose, and Moore--Penrose pseudoinverse. \\
$\preceq,\succeq$ & Positive-semidefinite matrix ordering. \\
$\|\cdot\|_2,\langle\cdot,\cdot\rangle$ & Euclidean norm and inner product for vectors. \\
\end{longtable}
\addtocounter{table}{-1}
\endgroup

\section{LMMSE-estimator and the conditional covariance}

Consider a linear observation model with Gaussian noise
\begin{equation}
\label{eq:app-obs-model}
\bm{v}
  = \mathbf{B}\bm{u} + \sigma\,\bm{n},
\qquad
\bm{n}\sim\mathcal{N}(\mathbf{0},\mathbf{I}),
\qquad
\mathbb{E}[\bm{u}]=\mathbb{E}[\bm{v}]=\mathbf{0},
\end{equation}
with $\mathrm{Cov}(\bm{u},\bm{n})=\mathbf{0}$.  
For random vectors $\bm{p},\bm{q}$ we write
\[
\mathrm{Cov}(\bm{p})=\mathbb{E}[\bm{p}\bm{p}^{\!\mathsf{T}}],
\qquad
\mathrm{Cov}(\bm{p},\bm{q})=\mathbb{E}[\bm{p}\bm{q}^{\!\mathsf{T}}].
\]
Let $\mathbf{\Sigma}_{\bm{u}}=\mathrm{Cov}(\bm{u})$.  
Linearity of expectations gives
\begin{equation}
\mathbf{\Sigma}_{\bm{v}}
    = \mathbf{B}\mathbf{\Sigma}_{\bm{u}}\mathbf{B}^{\!\mathsf{T}}
      + \sigma^2\mathbf{I},
\qquad
\mathbf{\Sigma}_{\bm{u}\bm{v}}
    = \mathrm{Cov}(\bm{u},\bm{v})
    = \mathbf{\Sigma}_{\bm{u}}\mathbf{B}^{\!\mathsf{T}}.
\end{equation}

The LMMSE estimator \cite{kay1993fundamentals} is the linear map $\mathbf{D}$ that minimizes
$\mathbb{E}\!\left[\lVert \bm{u}-\mathbf{D}\bm{v}\rVert^2\right]$.  
By the orthogonality principle,
\begin{equation}
\mathbf{D}
  = \mathbf{\Sigma}_{\bm{u}\bm{v}}\,
    \mathbf{\Sigma}_{\bm{v}}^{\dagger}.
\end{equation}
The corresponding error covariance is
\begin{equation}
\label{eq:appendix-error-covariance}
\mathbf{\Sigma}_{\bm{u}\mid\bm{v}}
   = \mathrm{Cov}(\bm{u}-\hat{\bm{u}})
   = \mathbf{\Sigma}_{\bm{u}}
     - \mathbf{\Sigma}_{\bm{u}\bm{v}}\,
       \mathbf{\Sigma}_{\bm{v}}^{\dagger}\,
       \mathbf{\Sigma}_{\bm{v}\bm{u}}.
\end{equation}
This matrix quantifies the reduction in uncertainty about $\bm{u}$ after observing $\bm{v}$.  
It is positive semidefinite $\mathbf{\Sigma}_{\bm{u}\mid\bm{v}} \succeq 0$ and 
$\mathbf{\Sigma}_{\bm{u}\mid\bm{v}} \preceq \mathbf{\Sigma}_{\bm{u}}$,  
making it a natural measure of estimator quality. Smaller values indicate a more informative observation and a more accurate estimation.  
For an identity observation $\mathbf{B}=\mathbf{I}$,  
$\mathbf{\Sigma}_{\bm{u}\mid\bm{v}}$ reduces directly to the posterior map $\Phi(\mathbf{\Sigma}_{\bm{u}};\sigma)$ used in the main text
 in \eqref{eq:posterior-map}.

\subsection{Conditional covariance $\mathbf{\Sigma}_{H\mid L}$}

We now apply the LMMSE framework to quantify the uncertainty in the high-frequency content of a signal after its low-frequency content is known.  
Let $\bm{u}$ be a zero-mean random vector, and let $\bm{u}_L$ and $\bm{u}_H$ denote the components of $\bm{u}$ associated with the low- and high-frequency subspaces, respectively.  
Viewing $\bm{u}_H$ as the unknown quantity and $\bm{u}_L$ as an available noise-less observation, we may treat $\bm{u}_L$ as the vector $\bm{v}$ in the LMMSE model (\eqref{eq:app-obs-model}) with observation matrix $\mathbf{B}$ equal to the selector that extracts the LF components.  
The LMMSE error covariance (\eqref{eq:appendix-error-covariance}) directly gives the conditional HF covariance
\begin{equation}
\mathbf{\Sigma}_{H\mid L}
   = \mathbf{\Sigma}_H
     - \mathbf{\Sigma}_{HL}\,\mathbf{\Sigma}_L^\dagger\,\mathbf{\Sigma}_{LH},
\end{equation}
where $\mathbf{\Sigma}_H=\mathrm{Cov}(\bm{u}_H)$, $\mathbf{\Sigma}_L=\mathrm{Cov}(\bm{u}_L)$, and $\mathbf{\Sigma}_{HL}=\mathrm{Cov}(\bm{u}_H,\bm{u}_L)$.  
This matrix characterizes the uncertainty of the HF component when LF information is known.  When the LF and HF components are uncorrelated ($\mathbf{\Sigma}_{HL}=0$), the conditional covariance reduces to $\mathbf{\Sigma}_{H\mid L}=\mathbf{\Sigma}_H$, indicating that the LF part carries no predictive information about the HF part.

\section{Proof of Theorem \ref{thm:sigmamax} (Start bound)}
\label{app:thm-sigmamax}
Recall that
\[
\bm{\Delta}_t^{\max}
=
\bm{\Sigma}_{H\mid L}
-
\Phi(\bm{\Sigma}_{H\mid L};\sigma_t)
=
\bm{\Sigma}_{H\mid L}
(\bm{\Sigma}_{H\mid L}+\sigma_t^2\mathbf{I})^{-1}
\bm{\Sigma}_{H\mid L},
\]
where we used the posterior identity
\[
\Phi(\bm{\Sigma};\sigma)
=
\bm{\Sigma}
-
\bm{\Sigma}(\bm{\Sigma}+\sigma^2\mathbf{I})^{-1}\bm{\Sigma}.
\]
For brevity, write
\[
\bm{\Sigma} := \bm{\Sigma}_{H\mid L},
\qquad
\bm{\Delta} := \bm{\Delta}_t^{\max}.
\]

\paragraph{Eigen-decomposition and diagonalization.}
Let
\begin{equation}
\bm{\Sigma}
= \bm{q}\,\Lambda\,\bm{q}^{\!\mathsf{T}},
\qquad
\Lambda = \mathrm{diag}(\nu_1,\dots,\nu_D),
\qquad
\nu_i \ge 0.
\end{equation}
Because
\[
\bm{\Sigma} + \sigma_t^2 \mathbf{I}
= \bm{q}(\Lambda + \sigma_t^2\mathbf{I})\bm{q}^{\!\mathsf{T}},
\qquad
(\bm{\Sigma} + \sigma_t^2 \mathbf{I})^{-1}
= \bm{q}(\Lambda + \sigma_t^2\mathbf{I})^{-1}\bm{q}^{\!\mathsf{T}},
\]
we obtain
\begin{align}
\bm{\Delta}
&=
\bm{\Sigma}(\bm{\Sigma}+\sigma_t^2\mathbf{I})^{-1}\bm{\Sigma} \\
&=
\bm{q}\,\Lambda(\Lambda + \sigma_t^2\mathbf{I})^{-1}\Lambda\,\bm{q}^{\!\mathsf{T}}.
\end{align}
Thus \(\bm{\Sigma}\) and \(\bm{\Delta}\) are simultaneously diagonalizable with eigenvectors \(\bm{q}\), and the eigenvalues of \(\bm{\Delta}\) are
\begin{equation}
\frac{\nu_i^2}{\nu_i + \sigma_t^2},
\qquad i = 1,\dots,D.
\end{equation}

\paragraph{Matrix inequality as scalar inequalities.}
The condition
\[
\bm{\Delta} \preceq \tau_{\max}\,\bm{\Sigma}
\]
is equivalent, in the common eigenbasis, to
\begin{equation}
\frac{\nu_i^2}{\nu_i + \sigma_t^2}
\;\le\;
\tau_{\max}\,\nu_i
\qquad\text{for all eigenvalues }\nu_i>0.
\label{eq:scalarineq}
\end{equation}

\paragraph{Solving the scalar inequality.}
Fix any \(\nu>0\).  
Inequality~\eqref{eq:scalarineq} becomes
\[
\frac{\nu^2}{\nu+\sigma_t^2} \le \tau_{\max}\nu.
\]
Cancelling the factor \(\nu>0\) gives
\[
\frac{\nu}{\nu+\sigma_t^2} \le \tau_{\max}.
\]
Multiply both sides by \(\nu+\sigma_t^2\):
\[
\nu \;\le\; \tau_{\max}(\nu+\sigma_t^2)
= \tau_{\max}\nu + \tau_{\max}\sigma_t^2.
\]
Rearranging yields
\[
(1-\tau_{\max})\,\nu \;\le\; \tau_{\max}\,\sigma_t^2.
\]
Since \(\tau_{\max}>0\),
\[
\sigma_t^2 \;\ge\; \frac{1-\tau_{\max}}{\tau_{\max}}\,\nu.
\]

\paragraph{Worst-case mode and conclusion.}
The bound must hold for every eigenvalue \(\nu_i\).  
Because the right-hand side increases with \(\nu\), the most restrictive condition occurs at the largest eigenvalue \(\nu_{\max}(\bm{\Sigma}_{H\mid L})\).  
Thus
\[
\sigma_t^2 \;\ge\;
\frac{1-\tau_{\max}}{\tau_{\max}}\,
\nu_{\max}(\bm{\Sigma}_{H\mid L}).
\]
Identifying the minimal feasible starting noise level with \(\sigma_{\max}\) proves the claim.

\section{Proof of Corollary \ref{cor:sigmamax} (WSS/PSD assumption)}
\label{app:cor-sigmamax}
Under the WSS assumption, the covariance of $\bm{x}_0$ is diagonalized by the discrete Fourier transform (DFT) \cite{papoulis2002probability}.  
Let $\widehat{\bm{x}}_0 = \mathbf{F}\bm{x}_0$ and note that $\mathbf{F}$ is unitary.  
Then
\begin{equation}
\widehat{\bm{\Sigma}}_0
:= \mathrm{Cov}(\widehat{\bm{x}}_0)
= \mathbf{F}\,\bm{\Sigma}_0\,\mathbf{F}^{\!\mathsf{H}}
= \mathrm{diag}\bigl(S(\omega) : \omega \in \Omega \bigr),
\end{equation}
where $S(\omega)$ is the power spectral density (PSD).  
Because $\mathbf{F}$ is unitary, it preserves eigenvalues and Loewner order:
\[
\bm{A} \preceq \bm{B}
\quad\Longleftrightarrow\quad
\mathbf{F}\bm{A}\mathbf{F}^{\!\mathsf{H}}
\preceq
\mathbf{F}\bm{B}\mathbf{F}^{\!\mathsf{H}}.
\]
Thus transforming to the frequency domain does not change the eigenvalue bounds in Theorem~\ref{thm:sigmamax}; it simply diagonalizes the matrices involved.

\paragraph{LF/HF splitting under WSS.}
An ideal LF/HF mask corresponds to diagonal projection matrices $\mathbf{P}_L$ and $\mathbf{P}_H=\mathbf{I}-\mathbf{P}_L$ selecting disjoint frequency sets $\Omega_L$ and $\Omega_H$.  
Since $\widehat{\bm{\Sigma}}_0$ is diagonal, there is no cross-covariance between LF and HF components:
\begin{equation}
\widehat{\bm{\Sigma}}_{HL}
= \mathbf{P}_H \widehat{\bm{\Sigma}}_0 \mathbf{P}_L
= \mathbf{0}.
\end{equation}
Therefore the HF covariance is simply
\begin{equation}
\widehat{\bm{\Sigma}}_H
= \mathbf{P}_H \widehat{\bm{\Sigma}}_0 \mathbf{P}_H
= \mathrm{diag}\bigl(S(\omega): \omega \in \Omega_H\bigr).
\end{equation}

\paragraph{Conditional HF covariance.}
The conditional covariance of HF given LF satisfies
\[
\bm{\Sigma}_{H\mid L}
= \bm{\Sigma}_{H} - \bm{\Sigma}_{HL}\bm{\Sigma}_L^\dagger \bm{\Sigma}_{LH}.
\]
Since $\bm{\Sigma}_{HL}=0$ under WSS with an ideal LF mask, we have
\[
\bm{\Sigma}_{H\mid L}
= \widehat{\bm{\Sigma}}_{H\mid L}
= \widehat{\bm{\Sigma}}_H
= \mathrm{diag}\bigl(S(\omega): \omega\in\Omega_H\bigr).
\]
Thus the eigenvalues of $\bm{\Sigma}_{H\mid L}$ are precisely the PSD values $S(\omega)$ on the HF set $\Omega_H$.  
In particular,
\begin{equation}
\nu_{\max}(\bm{\Sigma}_{H\mid L})
= \max_{\omega\in\Omega_H} S(\omega).
\end{equation}

\paragraph{Substitution into the start bound.}
Theorem~\ref{thm:sigmamax} states that
\[
\sigma_{\max}^2
\;\ge\;
\frac{1-\tau_{\max}}{\tau_{\max}}\,
\nu_{\max}(\bm{\Sigma}_{H\mid L}).
\]
Substituting the PSD expression for $\nu_{\max}(\bm{\Sigma}_{H\mid L})$ yields
\[
\sigma_{\max}^2
\;\ge\;
\frac{1-\tau_{\max}}{\tau_{\max}}\,
\max_{\omega\in\Omega_H} S(\omega),
\]
which is exactly the claimed PSD form~\eqref{eq:sigmamax-psd}.

\section{Proof of Theorem \ref{thm:sigmamin} (Stop bound)}
\label{app:thm-sigmamin}
The proof follows the same diagonalization argument used in Theorem~\ref{thm:sigmamax}.  
Since the posterior map 
\(
\Phi(\bm{\Sigma};\sigma) 
\)
is the same functional form appearing in the start-bound proof, it is diagonalized by the eigenvectors of $\bm{\Sigma}$.  
Thus, as in the previous theorem, all Loewner inequalities may be checked mode-wise in the eigenbasis of $\bm{\Sigma}_{\mathbf{s}}$.

\paragraph{Eigenbasis and per-mode posterior variances.}
Let
\begin{equation}
\bm{\Sigma}_{\mathbf{s}}
  = \bm{q}\,\Lambda\,\bm{q}^{\!\mathsf{T}},
  \qquad
  \Lambda = \mathrm{diag}(\nu_1,\dots,\nu_D),
  \qquad
  \nu_i \ge 0.
\end{equation}
Since $\Phi(\bm{\Sigma};\sigma)$ is diagonalized by the same eigenvectors as $\bm{\Sigma}$ (as in the proof of Theorem~\ref{thm:sigmamax}),  
the matrices
\[
\bm{\Sigma}_0 = \Phi(\bm{\Sigma}_{\mathbf{s}};\sigma_{\mathbf{s}})
\quad\text{and}\quad
\Phi(\bm{\Sigma}_{\mathbf{s}}; \sqrt{\sigma^2_{\mathbf{s}}+\sigma^2_t})
\]
are diagonal in the same basis $\bm{q}$.  
Their diagonal entries are obtained by applying the scalar posterior formula
\[
\frac{\nu_i\,\sigma^2}{\nu_i+\sigma^2}
\]
to each eigenvalue $\nu_i$ of $\bm{\Sigma}_{\mathbf{s}}$.

Thus, per mode $\nu_i$,
\begin{align}
\text{eigenvalue of } \bm{\Sigma}_0 &: 
&& \frac{\nu_i\,\sigma_{\mathbf{s}}^2}{\nu_i+\sigma_{\mathbf{s}}^2},\\[3pt]
\text{eigenvalue of } \Phi(\bm{\Sigma}_{\mathbf{s}};\sqrt{\sigma_{\mathbf{s}}^2+\sigma_t^2}) &: 
&& \frac{\nu_i\,(\sigma_{\mathbf{s}}^2+\sigma_t^2)}{\nu_i+\sigma_{\mathbf{s}}^2+\sigma_t^2}.
\end{align}

\paragraph{Residual and tolerance condition.}
The residual covariance is
\[
\bm{\Delta}_t^{\min}
  =
  \Phi\!\left(\bm{\Sigma}_{\mathbf{s}};\sqrt{\sigma_{\mathbf{s}}^2+\sigma_t^2}\right)
  - \bm{\Sigma}_0
  \succeq \mathbf{0}.
\]
Since both matrices on the right-hand side share the eigenvectors of
$\bm{\Sigma}_{\mathbf{s}}$, the Loewner inequality
\[
\bm{\Delta}_t^{\min} \preceq \tau_{\min}\,\bm{\Sigma}_0
\]
is equivalent to a set of scalar inequalities on their eigenvalues.

Let $\nu>0$ be any eigenvalue of $\bm{\Sigma}_{\mathbf{s}}$.  
The corresponding eigenvalues of
$\bm{\Sigma}_0 = \Phi(\bm{\Sigma}_{\mathbf{s}};\sigma_{\mathbf{s}})$
and
$\Phi(\bm{\Sigma}_{\mathbf{s}};\sqrt{\sigma_{\mathbf{s}}^2+\sigma_t^2})$
are, respectively,
\[
\frac{\nu\,\sigma_{\mathbf{s}}^2}{\nu+\sigma_{\mathbf{s}}^2}
\qquad\text{and}\qquad
\frac{\nu\,(\sigma_{\mathbf{s}}^2+\sigma_t^2)}
     {\nu+\sigma_{\mathbf{s}}^2+\sigma_t^2}.
\]
Thus the matrix inequality
\(
\bm{\Delta}_t^{\min} \preceq \tau_{\min}\,\bm{\Sigma}_0
\)
is equivalent to requiring, for every $\nu>0$,
\begin{equation}
\frac{\nu\,(\sigma_{\mathbf{s}}^2+\sigma_t^2)}
     {\nu+\sigma_{\mathbf{s}}^2+\sigma_t^2}
-
\frac{\nu\,\sigma_{\mathbf{s}}^2}{\nu+\sigma_{\mathbf{s}}^2}
\;\le\;
\tau_{\min}\,
\frac{\nu\,\sigma_{\mathbf{s}}^2}{\nu+\sigma_{\mathbf{s}}^2}.
\end{equation}

Canceling the common factor $\nu>0$ gives
\begin{equation}
\frac{\sigma_{\mathbf{s}}^2+\sigma_t^2}{\nu+\sigma_{\mathbf{s}}^2+\sigma_t^2} -
\frac{\sigma_{\mathbf{s}}^2}{\nu+\sigma_{\mathbf{s}}^2}
\;\le\;
\tau_{\min}\,
\frac{\sigma_{\mathbf{s}}^2}{\nu+\sigma_{\mathbf{s}}^2}.
\end{equation}

Multiplying both sides by
$(\nu+\sigma_{\mathbf{s}}^2)(\nu+\sigma_{\mathbf{s}}^2+\sigma_t^2)$
leads to the linear inequality in $\sigma_t^2$:
\begin{equation}
(\nu - \tau_{\min}\sigma_{\mathbf{s}}^2)\,\sigma_t^2
\;\le\;
\tau_{\min}\,\sigma_{\mathbf{s}}^2\,(\nu+\sigma_{\mathbf{s}}^2).
\label{eq:sigmamin-linear-2}
\end{equation}

\paragraph{Active and inactive modes.}
If $\nu \le \tau_{\min}\sigma_{\mathbf{s}}^2$, then the left-hand side of 
\eqref{eq:sigmamin-linear-2} is non-positive and the right-hand side is non-negative;  
the inequality holds automatically for all $\sigma_t^2\ge 0$.  
Such modes impose no constraint.

If $\nu > \tau_{\min}\sigma_{\mathbf{s}}^2$, we divide by the positive factor 
$\nu - \tau_{\min}\sigma_{\mathbf{s}}^2$ to obtain
\begin{equation}
\sigma_t^2
  \;\le\;
  \frac{\tau_{\min}\,\sigma_{\mathbf{s}}^2\,(\nu+\sigma_{\mathbf{s}}^2)}
       {\nu-\tau_{\min}\sigma_{\mathbf{s}}^2}.
\label{eq:sigmamin-per-mode-final}
\end{equation}

\paragraph{Worst-case eigenvalue and final bound.}
The right-hand side of \eqref{eq:sigmamin-per-mode-final} is a strictly decreasing function of $\nu$ on $(\tau_{\min}\sigma_{\mathbf{s}}^2,\infty)$.  
Thus the most restrictive constraint is obtained by taking the largest eigenvalue 
$\nu_{\max}=\nu_{\max}(\bm{\Sigma}_{\mathbf{s}})$, giving
\begin{equation}
\sigma_t^2
  \;\le\;
  \frac{\tau_{\min}\,\sigma_{\mathbf{s}}^2\,(\nu_{\max}+\sigma_{\mathbf{s}}^2)}
       {\nu_{\max}-\tau_{\min}\sigma_{\mathbf{s}}^2}.
\end{equation}
If $\nu_{\max} \le \tau_{\min}\sigma_{\mathbf{s}}^2$, then all modes are inactive and the constraint is vacuous.

Identifying the maximal allowable reverse-diffusion variance at termination with $\sigma_{\min}^2$ produces
\[
\sigma_{\min}^2 \;\le\; \frac{\tau_{\min}\,\sigma_{\mathbf{s}}^2\,(\nu_{\max}+\sigma_{\mathbf{s}}^2)}
       {\nu_{\max}-\tau_{\min}\sigma_{\mathbf{s}}^2},
\]
which establishes the claim.

\section{Proof of Corollary \ref{cor:sigmamin} (High-SNR assumption)}
\label{app:cor-sigmamin}
From Theorem~\ref{thm:sigmamin}, the upper bound on $\sigma_{\min}^2$ is
\begin{equation}
\frac{\tau_{\min}\,\sigma_{\mathbf{s}}^2\,(\nu_{\max}+\sigma_{\mathbf{s}}^2)}
     {\nu_{\max}-\tau_{\min}\sigma_{\mathbf{s}}^2},
\end{equation}
where $\nu_{\max}=\nu_{\max}(\bm{\Sigma}_{\mathbf{s}})$ is the largest eigenvalue of the clean-signal covariance.
Factor out $\nu_{\max}$ from numerator and denominator:
\begin{equation}
\tau_{\min}\,\sigma_{\mathbf{s}}^2\,
\frac{\nu_{\max}+\sigma_{\mathbf{s}}^2}{\nu_{\max}-\tau_{\min}\sigma_{\mathbf{s}}^2}
=
\tau_{\min}\,\sigma_{\mathbf{s}}^2\,
\frac{1+\sigma_{\mathbf{s}}^2/\nu_{\max}}{1-\tau_{\min}\sigma_{\mathbf{s}}^2/\nu_{\max}}.
\end{equation}
In the high-SNR regime $\nu_{\max}\gg\sigma_{\mathbf{s}}^2$, both ratios $\sigma_{\mathbf{s}}^2/\nu_{\max}$ and $\tau_{\min}\sigma_{\mathbf{s}}^2/\nu_{\max}$ are small, and we obtain the approximation
\begin{equation}
\frac{1+\sigma_{\mathbf{s}}^2/\nu_{\max}}{1-\tau_{\min}\sigma_{\mathbf{s}}^2/\nu_{\max}}
\;\approx\;
1,
\end{equation}
so that
\begin{equation}
\frac{\tau_{\min}\,\sigma_{\mathbf{s}}^2\,(\nu_{\max}+\sigma_{\mathbf{s}}^2)}
     {\nu_{\max}-\tau_{\min}\sigma_{\mathbf{s}}^2}
\;\approx\;
\tau_{\min}\,\sigma_{\mathbf{s}}^2.
\end{equation}
Then, from Theorem~\ref{thm:sigmamin}, we conclude that in this regime
\begin{equation}
\sigma_{\min}^2 \;\lesssim\; \tau_{\min}\,\sigma_{\mathbf{s}}^2,
\end{equation}
which is the claim.

\section{Proof of Theorem \ref{thm:invariance} (Re-scaling invariance)}
\label{app:thm-invariance}
We consider the unit-gradient update form used in~\eqref{eq:ugd}.  
At iteration $k$, the update applied to $(f,g)$ is
\begin{equation}
\Delta\bm{\mu}^{k}
  = -\hat{\alpha}^k \!\left(
      \frac{\nabla f^k}{\|\nabla f^k\|_2}
      +\hat{\lambda}\,\frac{\nabla g^k}{\|\nabla g^k\|_2}
    \right),
\label{eq:ugd-again}
\end{equation}
where $\hat{\alpha}^k>0$ and $\hat{\lambda}>0$ is a global sampler parameters.

Let $\phi_f,\phi_g:\mathbb{R}\to\mathbb{R}$ be strictly increasing $C^1$ functions and define
\[
\widetilde f = \phi_f \!\circ f,
\qquad
\widetilde g = \phi_g \!\circ g.
\]
By the chain rule,
\begin{equation}
\nabla \widetilde f^k = \phi_f'(f^k)\,\nabla f^k,
\qquad
\nabla \widetilde g^k = \phi_g'(g^k)\,\nabla g^k,
\end{equation}
with $\phi_f'(f^k)>0$ and $\phi_g'(g^k)>0$ because $\phi_f,\phi_g$ are strictly increasing.  
Hence
\begin{equation}
\|\nabla \widetilde f^k\|_2
  = \phi_f'(f^k)\,\|\nabla f^k\|_2,
\qquad
\|\nabla \widetilde g^k\|_2
  = \phi_g'(g^k)\,\|\nabla g^k\|_2.
\end{equation}

Therefore the normalized gradients are unchanged:
\begin{equation}
\frac{\nabla \widetilde f^k}{\|\nabla \widetilde f^k\|_2}
  = \frac{\phi_f'(f^k)\nabla f^k}{\phi_f'(f^k)\|\nabla f^k\|_2}
  = \frac{\nabla f^k}{\|\nabla f^k\|_2},
\end{equation}
\begin{equation}
\frac{\nabla \widetilde g^k}{\|\nabla \widetilde g^k\|_2}
  = \frac{\phi_g'(g^k)\nabla g^k}{\phi_g'(g^k)\|\nabla g^k\|_2}
  = \frac{\nabla g^k}{\|\nabla g^k\|_2}.
\end{equation}

Applying the same unit-gradient update \eqref{eq:ugd-again} to $(\widetilde f,\widetilde g)$ with the same parameters $(\hat{\alpha}^k,\hat{\lambda})$ yields
\[
\Delta\widetilde{\bm{\mu}}^{k}
  = -\hat{\alpha}^k\!\left(
      \frac{\nabla \widetilde f^k}{\|\nabla \widetilde f^k\|_2}
      +\hat{\lambda}\,\frac{\nabla \widetilde g^k}{\|\nabla \widetilde g^k\|_2}
    \right)
  = -\hat{\alpha}^k\!\left(
      \frac{\nabla f^k}{\|\nabla f^k\|_2}
      +\hat{\lambda}\,\frac{\nabla g^k}{\|\nabla g^k\|_2}
    \right)
  = \Delta\bm{\mu}^{k}.
\]

Thus both the direction and the magnitude of the update are invariant under any strictly increasing $C^1$ reparameterizations of $f$ and $g$ (including positive scalings).

\section{Proof of Theorem \ref{thm:descent} (Common descent condition)}
\label{app:thm-descent}
Let
\[
d^k = \nabla f^k + \lambda^k \nabla g^k,
\qquad
\cos\vartheta^k
= \frac{\langle \nabla f^k, \nabla g^k\rangle}
       {\|\nabla f^k\|_2\,\|\nabla g^k\|_2},
\qquad
r^k = \frac{\|\nabla f^k\|_2}{\|\nabla g^k\|_2}.
\]
Simultaneous descent \cite{fliege2000steepest} requires
\[
\langle \nabla f^k, d^k\rangle > 0
\quad\text{and}\quad
\langle \nabla g^k, d^k\rangle > 0.
\]

We compute both inner products in terms of
\(\cos\vartheta^k\), \(r^k\), and \(\lambda^k\).  
First,
\begin{align}
\langle \nabla f^k, d^k\rangle
&= \|\nabla f^k\|_2^2 
   + \lambda^k \langle \nabla f^k, \nabla g^k\rangle \\
&= \|\nabla f^k\|_2^2
   + \lambda^k \|\nabla f^k\|_2\|\nabla g^k\|_2 \cos\vartheta^k \\
&= \|\nabla f^k\|_2^2
   \bigl( 1 + \lambda^k \tfrac{\cos\vartheta^k}{r^k} \bigr).
\label{eq:descent-f-clean}
\end{align}
Similarly,
\begin{align}
\langle \nabla g^k, d^k\rangle
&= \langle \nabla g^k, \nabla f^k\rangle
   + \lambda^k \|\nabla g^k\|_2^2 \\
&= \|\nabla g^k\|_2^2 \bigl( r^k\cos\vartheta^k + \lambda^k \bigr).
\label{eq:descent-g-clean}
\end{align}
Since both norm-squared factors are strictly positive, the signs of
\eqref{eq:descent-f-clean}–\eqref{eq:descent-g-clean} are determined by
\[
1 + \lambda^k \frac{\cos\vartheta^k}{r^k}
\quad\text{and}\quad
r^k\cos\vartheta^k + \lambda^k.
\]

We now distinguish two geometric regimes.

\paragraph{Case 1: $\cos\vartheta^k \ge 0$.}
The gradients form an acute angle.  
Since $\lambda^k > 0$ by construction, and noting that $\cos\vartheta^k\ge0$, $r^k>0$, we obtain  
\[
1 + \lambda^k \frac{\cos\vartheta^k}{r^k} \ge 1 > 0,
\qquad
r^k\cos\vartheta^k + \lambda^k \ge 0,
\]
with strict positivity whenever at least one gradient is nonzero (as assumed). 
Thus every \(\lambda^k > 0\) yields simultaneous descent.

\paragraph{Case 2: $\cos\vartheta^k < 0$.}
The gradients form an obtuse angle; the two inequalities now impose nontrivial constraints.

From \eqref{eq:descent-f-clean},
\[
1 + \lambda^k \frac{\cos\vartheta^k}{r^k} > 0
\quad\Longleftrightarrow\quad
\lambda^k < -\,\frac{r^k}{\cos\vartheta^k}.
\]
Here $\cos\vartheta^k<0$ and $r^k>0$, so multiplying by $r^k/\cos\vartheta^k$ reverses the inequality.

From \eqref{eq:descent-g-clean},
\[
r^k\cos\vartheta^k + \lambda^k > 0
\quad\Longleftrightarrow\quad
\lambda^k > -\,r^k\cos\vartheta^k.
\]

Combining both constraints gives the feasible interval
\[
\lambda^k \in 
\bigl( r^k(-\cos\vartheta^k),\; r^k(-1/\cos\vartheta^k) \bigr),
\]
which is exactly the claimed condition for $\cos\vartheta^k < 0$.
\textbf{Note.} In the degenerate edge case where $cos^k=-1$ (perfect anti-alignment), the admissible $\hat\lambda$ value would be 1, and there is no update to $\mu^k$. Practically, we never observe $cos^k=-1$ due to numerical computation.

\section{Greedy-optimal sampler hyperparameters}
\label{app:opt-weights}
We detail the update rules and the optimal-hyperparameters scheme for the suite of optimizers tested in this work.
The proposed scheme resembles approaches in the compressed sensing and diffusion-based reconstruction literature, where unbiased MSE estimators have been employed for hyperparameter tuning \cite{iyer2020sure, weller2014monte, ozturkler2023smrd}.

\begin{equation}
    \bm{\mu}^{k+1}
    \leftarrow \bm{\mu}^k
    - \bm{U}^k \, \hat{\bm{w}}^{k},
    \qquad
    \hat{\bm{w}}^k
    = \argmin_{\bm{w} \in \mathbb{R}^2_+}
    \left\| \bm{x}^* -
    \left( \bm{\mu}^k - \bm{U}^k \bm{w} \right) \right\|_2^2.
\end{equation}

\paragraph{Vanilla}
The first-order vanilla update is
\begin{align}
    \bm{v}^{k+1}
    &\leftarrow \alpha^k \left( \nabla f^k + \lambda^k \nabla g^k \right), \\
    \bm{\mu}^{k+1}
    &\leftarrow \bm{\mu}^k - \bm{v}^{k+1}.
\end{align}

These updates yield
\begin{equation}
    \bm{U}^k =
    \begin{bmatrix}
        \nabla f^k & \nabla g^k
    \end{bmatrix},
    \qquad
    \bm{w}^k :=
    \begin{bmatrix}
        \alpha^k \\
        \alpha^k \lambda^k
    \end{bmatrix}.
\end{equation}

\paragraph{Momentum}
The optimal hyperparameters scheme becomes: 

\begin{equation}
    \hat{\bm{w}}^k
    = \argmin_{\bm{w} \in \mathbb{R}^3_+}
    \left\| \bm{x}^* -
    \left( \bm{\mu}^k - \bm{U}^k \bm{w} \right) \right\|_2^2.
\end{equation}

The momentum update becomes
\begin{align}
    \bm{v}^{k+1}
    &\leftarrow \alpha^k \left( \nabla f^k + \lambda^k \nabla g^k \right)
        + \beta^k \bm{v}^k, \\
    \bm{\mu}^{k+1}
    &\leftarrow \bm{\mu}^k - \bm{v}^{k+1}.
\end{align}

These updates yield
\begin{equation}
    \bm{U}^k =
    \begin{bmatrix}
        \nabla f^k & \nabla g^k & \bm{v}^k
    \end{bmatrix},
    \qquad
    \bm{w}^k :=
    \begin{bmatrix}
        \alpha^k \\
        \alpha^k \lambda^k \\
        \beta^k
    \end{bmatrix}.
\end{equation}

\paragraph{Momentum + Preconditioner}
Adding a polynomial preconditioner for linear measurement operator $\mathbf{A}$,
\begin{align}
    \bm{v}^{k+1}
    &\leftarrow \alpha^k \left(
        \text{pol}(\mathbf{A}^\mathsf{T}\mathbf{A}) \nabla f^k
        + \lambda^k \nabla g^k
    \right)
    + \beta^k \bm{v}^k, \\
    \bm{\mu}^{k+1}
    &\leftarrow \bm{\mu}^k - \bm{v}^{k+1}.
\end{align}

These updates yield
\begin{equation}
    \bm{U}^k =
    \begin{bmatrix}
        \text{pol}(\mathbf{A}^\mathsf{T}\mathbf{A}) \nabla f^k
        &
        \nabla g^k
        &
        \bm{v}^k
    \end{bmatrix},
    \qquad
    \bm{w}^k :=
    \begin{bmatrix}
        \alpha^k \\
        \alpha^k \lambda^k \\
        \beta^k
    \end{bmatrix}.
\end{equation}

\paragraph{Sensitivity to tuning-set size.}
We evaluate the sensitivity of $\hat{\lambda}$ to tuning-set size and image selection using the ImageNet Gaussian deblurring task. For tuning-set sizes of 2, 3, 5, and 10 images, we randomly sample 20 sets of images for each size and independently tune $\hat{\lambda}$ for each set. Table~\ref{tab:tuning_sensitivity} reports the mean and standard deviation of the resulting values across the 20 sets. The mean remains stable across tuning-set sizes, ranging from 0.939 to 0.947, while the standard deviation decreases from 0.030 to 0.013 as the tuning-set size increases. These results suggest that $\hat{\lambda}$ can be reliably selected using a small tuning set, with limited sensitivity to image selection.

\begin{table}[h]
    \centering
    \caption{Sensitivity analysis of tuning-set size on ImageNet Gaussian deblurring.}
    \label{tab:tuning_sensitivity}
    \begin{tabular}{cc}
        \toprule
        \textbf{Tuning-set size} & \textbf{$\hat{\lambda}$ values} \\
        \midrule
        2  & $0.947 \pm 0.030$ \\
        3  & $0.944 \pm 0.023$ \\
        5  & $0.945 \pm 0.015$ \\
        10 & $0.939 \pm 0.013$ \\
        \bottomrule
    \end{tabular}
\end{table}

\section{Datasets and Implementation Details}
\label{app:impl-details}
We implemented the OptDiff pipeline with  momentum and preconditioner addons in our GitHub repository. The repository is publicly available. Single-image reconstruction runtimes were obtained on a single NVIDIA A40 GPU with 48 GB memory.

\subsection{Preconditioners}
We evaluated several polynomial preconditioners from the literature \cite{johnson1983polynomial, zulfiquar2015improved, iyer2024polynomial}, with polynomial degrees ranging from 1 to 10. The best performance was achieved by the degree-2 preconditioner proposed by \citet{iyer2024polynomial}.

\subsection{MRI Reconstruction} 
\paragraph{Datasets:} We use the fastMRI T2-weighted dataset \cite{knoll2020fastmri} with the following splits:
\begin{itemize}
    \item Tuning set, 10 slices from the validation dataset used for hyperparameter tuning and invariance experiments (Subsection \ref{sub:invariance})
    \item Ablation set, 275 slices from the validation dataset used for the ablation experiments (Section \ref{sec:ablations})
    \item Test set, 330 slices from the test dataset used for comparisons against baseline methods (Subsection \ref{sub:comparisons})
\end{itemize}

\paragraph{Score model: } Score network from \citet{daras2023ambient}

\paragraph{Implementation: } In MRI, the measurement operator $\mathcal{A}$ is a complex valued mapping that incorporates coil sensitivities, Fourier Transform and sub-sampling operations. We implemented the MRI forward model and subsampling operations in our OptDiff repository. T2-weighted brain MR images from the fastMRI dataset were center cropped to $384\times384$ matrix size similar to \citet{jalal2021robust}. We applied coil compression to retain 8 virtual channels per slice. Coil sensitivity maps were estimated using ESPIRiT \cite{uecker2014espirit} algorithm.

Experiments were conducted with several configurations combining different acceleration factors $R=4$ and $R=8$, with calibration region sizes of $32$ and $16$ respectively, under both uniform and random undersampling masks. To estimate the effective noise floor, we used image-domain references reconstructed from fully sampled multicoil k-space data. After applying the same intensity normalization as in our experiments, we computed the root-mean-square intensity of background pixels outside the anatomy in the tuning set, following standard MRI noise-estimation procedures \cite{aja2009noise}. The procedure yielded a noise floor of $\sigma_{\mathbf{s}} = 0.021$.

\paragraph{Baseline methods:} 
We adapted the DPS, Pi-GDM, RED-diff and DAPS solvers in our codebase for baseline comparisons. We implemented DiffPIR \cite{zhu2023denoising} based on the original repository. For DDS, we used the original codebase \cite{chung2024decomposed}. We tuned the hyperparameters of all baseline methods for each setting. DDRM \cite{kawar2022denoising} and DDNM \cite{wang2022zero} require access to an explicit SVD of the measurement operator, which is computationally intractable for the multi-coil MRI forward model; therefore, they are not included as baselines for MRI reconstruction.

\subsection{Natural Image Tasks}

\paragraph{Datasets:} We use a subset of the ImageNet validation dataset (5 images for hyperparameter tuning, 330 images for testing) obtained from \href{https://www.kaggle.com/datasets/arjunashok33/miniimagenet?resource=download}{Kaggle}.
We use a subset of the FFHQ validation dataset (5 images for hyperparameter tuning, 330 images for testing).

\paragraph{Score models: } We adopt the unconditional pretrained diffusion models from \citet{mardani2023variational} and \citet{choi2021ilvr} for ImageNet and FFHQ, respectively.

\subsubsection{Degredations}

\paragraph{Deblurring: } As pre-processing, the images were cropped to $3 \times 256 \times 256$. The deblurring forward model was adopted from \citet{mardani2023variational}. We used kernel sizes $6, 12$ both with $\sigma=3$ and additional measurement noise with variance $0.01$ across both datasets.

\paragraph{Super-resolution: } As pre-processing, the images were cropped to $3 \times 256 \times 256$. For the experiments, each image is downsampled by a factor of $4, 16$ in both dimensions and corrupted with Gaussian noise. We used noise with variance $0.01$ in FFHQ and $0.02$ in ImageNet. The super-resolution forward model was adopted from \citet{mardani2023variational}.

\paragraph{Inpainting: } As pre-processing, the images were cropped to $3 \times 256 \times 256$. Each image was masked by a $64 \times 64$ square mask with randomly selected center coordinates (which was fixed across images). We added Gaussian noise with variance $0.01$ in FFHQ and $0.02$ in ImageNet. The masking forward model was adopted from \citet{mardani2023variational}.

\subsubsection{Baseline methods:} DPS, DDRM, Pi-GDM, RED-diff baselines are based on the repository from \citet{mardani2023variational}. DDNM and DiffPIR have been adapted into the \citet{mardani2023variational} repository. For DAPS, SITCOM and DMPlug, we used the original implementations from \citet{zhang2025improving,alkhouri2025sitcom,wang2024dmplug}, respectively. We used the hyperparameters provided by each repository as reference and further tuned them for each setting.

\section{Extended Limitations}
\label{app:limitations}
First, our noise-schedule design leverages forward models where the measurements preserve sufficient low-frequency information, enabling a reduced $\sigma_{\max}$. This reduction is available for the main tasks considered herein, including MRI reconstruction, deblurring, and super-resolution, but may be limited in problems such as inpainting, where the effective $\sigma_{\max}$ can match the default unconditional-generation value. Thus, the proposed design remains broadly applicable, but may not always reduce the starting noise level. Future work could extend the design beyond the endpoints ($\sigma_{\max}$ and $\sigma_{\min}$) toward principled rules for the curvature of the noise schedule, which determines how sampling steps are allocated across noise levels and, consequently, frequency regions.

Second, our sampler hyperparameter reparameterization builds upon RED-diff, an optimization-based posterior solver. Consequently, OptDiff inherits some limitations of this class of methods, which produce a single reconstruction and may favor pixelwise fidelity over perceptual quality, particularly in highly ill-posed settings with multiple plausible solutions \cite{daras2024survey,blau2018perception}. This limitation is most evident in $16\times$ super-resolution and inpainting, where sampling-based methods such as DAPS \cite{zhang2025improving} achieve better perceptual results, as shown in Table~\ref{tab:results-100steps}. Moreover, our NNLS-based hyperparameter tuning may reinforce this tendency, since its reference values are obtained by minimizing MSE. Future work could investigate more general tuning objectives that incorporate perceptual criteria, such as deep-feature losses, to better balance pixelwise distortion and perceptual similarity.

Lastly, the proposed reparameterizations, including the noise-schedule hyperparameters, are readily applicable to other posterior samplers (Subsection~\ref{sub:comparisons}, Table~\ref{tab:noise-sched-gen}). We note that the noise schedule design could be carried over to flow-based solvers through a change of variables, left as future work. Integrating our sampler hyperparameter design is less direct for methods that are not formulated as optimization problems. Extending this component of the framework to other posterior samplers would require additional theoretical justification, which we leave for future work.



\section{Comparison with baselines: full tables}

\begin{table}[h]
\centering
\caption{
Comparison with baselines using NFE-20. Our method consistently achieves the best or comparable performance across task--dataset pairs. Best results are shown in bold, and second-best results are underlined.}
\label{tab:results-20steps}
\resizebox{\linewidth}{!}{%
\begin{tabular}{c l ccc ccc}

\toprule
& & \multicolumn{6}{c}{\textbf{fastMRI}} \\
& & \multicolumn{3}{c}{\textbf{Acceleration R=4}} & \multicolumn{3}{c}{\textbf{Acceleration R=8}} \\
\cmidrule(lr){3-5} \cmidrule(lr){6-8}
\textbf{Task} & \textbf{Method}
& \textbf{PSNR} $\uparrow$ & \textbf{SSIM} $\uparrow$ & \cellcolor{gray!20}{}
& \textbf{PSNR} $\uparrow$ & \textbf{SSIM} $\uparrow$ & \cellcolor{gray!20}{} \\
\midrule

\multirow{7}{*}{\shortstack{MRI recon}}

& DPS \cite{chung2022diffusion}
& $31.93 \pm 1.50$ & $0.763 \pm 0.059$ & \cellcolor{gray!20}{}
& $27.36 \pm 1.48$ & $0.682 \pm 0.054$ & \cellcolor{gray!20}{} \\

& DiffPIR \cite{zhu2023denoising}
& $27.04 \pm 1.01$ & $0.727 \pm 0.054$ & \cellcolor{gray!20}{}
& $24.96 \pm 1.15$ & $0.587 \pm 0.075$ & \cellcolor{gray!20}{} \\

& Pi-GDM \cite{song2023pseudoinverse}
& $29.63 \pm 1.27$ & $0.739 \pm 0.063$ & \cellcolor{gray!20}{}
& $28.04 \pm 1.24$ & $0.707 \pm 0.062$ & \cellcolor{gray!20}{} \\

& RED-diff \cite{mardani2023variational}
& $33.24 \pm 1.49$ & $0.870 \pm 0.033$ & \cellcolor{gray!20}{}
& $29.43 \pm 1.58$ & $0.825 \pm 0.042$ & \cellcolor{gray!20}{} \\

& DDS \cite{chung2024decomposed}
& $30.27 \pm 1.27$ & $\underline{0.901 \pm 0.032}$ & \cellcolor{gray!20}{}
& $25.42 \pm 1.38$ & $0.765 \pm 0.031$ & \cellcolor{gray!20}{} \\

& DAPS \cite{zhang2025improving}
& $\underline{35.64 \pm 1.47}$ & $0.894 \pm 0.031$ & \cellcolor{gray!20}{}
& $\underline{30.82 \pm 1.45}$ & $\underline{0.848 \pm 0.036}$ & \cellcolor{gray!20}{} \\

\cmidrule(lr){2-8}

& OptDiff (Ours)
& $\mathbf{37.19 \pm 2.01}$ & $\mathbf{0.943 \pm 0.030}$ & \cellcolor{gray!20}{}
& $\mathbf{33.23 \pm 2.08}$ & $\mathbf{0.903 \pm 0.046}$ & \cellcolor{gray!20}{} \\

\bottomrule

\toprule
& & \multicolumn{3}{c}{\textbf{ImageNet}} & \multicolumn{3}{c}{\textbf{FFHQ}} \\
\cmidrule(lr){3-5} \cmidrule(lr){6-8}
\textbf{Task} & \textbf{Method}
& \textbf{PSNR} $\uparrow$ & \textbf{SSIM} $\uparrow$ & \textbf{LPIPS} $\downarrow$
& \textbf{PSNR} $\uparrow$ & \textbf{SSIM} $\uparrow$ & \textbf{LPIPS} $\downarrow$ \\
\cmidrule(lr){1-8}

\multirow{5}{*}{\shortstack{Deblurring \\ kernel $6$}}

& DDNM \cite{wang2022zero}
& $\mathbf{31.91 \pm 3.52}$ & $\mathbf{0.894 \pm 0.048}$ & $\mathbf{0.051 \pm 0.021}$
& $\mathbf{35.87 \pm 1.68}$ & $\mathbf{0.939 \pm 0.015}$ & $\underline{0.043 \pm 0.012}$ \\

& DiffPIR \cite{zhu2023denoising}
& $30.77 \pm 3.40$ & $0.853 \pm 0.051$ & $0.088 \pm 0.045$
& $\underline{35.11 \pm 1.72}$ & $0.918 \pm 0.015$ & $0.053 \pm 0.016$ \\

& RED-diff \cite{mardani2023variational} 
& $25.79 \pm 2.68$ & $0.677 \pm 0.074$ & $0.247 \pm 0.079$
& $29.07 \pm 1.15$ & $0.763 \pm 0.026$ & $0.147 \pm 0.033$ \\

& DAPS \cite{zhang2025improving}
& $29.67 \pm 1.97$ & $0.777 \pm 0.038$ & $0.148 \pm 0.046$
& $32.55 \pm 1.90$ & $0.875 \pm 0.021$ & $0.110 \pm 0.022$ \\

\cmidrule(lr){2-8}

& OptDiff (Ours)
& $\underline{31.66 \pm 3.81}$ & $\underline{0.890 \pm 0.047}$ & $\underline{0.082 \pm 0.038}$
& $\mathbf{35.87 \pm 1.72}$ & $\underline{0.937 \pm 0.014}$ & $\mathbf{0.039 \pm 0.011}$ \\

\midrule

\multirow{5}{*}{\shortstack{Severe deblur \\ kernel $12$}}

& DDNM \cite{wang2022zero}
& $\mathbf{27.89 \pm 3.87}$ & $\mathbf{0.782 \pm 0.110}$ & $\mathbf{0.136 \pm 0.056}$
& $\mathbf{32.33 \pm 2.06}$ & $\mathbf{0.890 \pm 0.030}$ & $\mathbf{0.080 \pm 0.022}$ \\

& DiffPIR \cite{zhu2023denoising}
& $\underline{27.27 \pm 3.73}$ & $\underline{0.739 \pm 0.122}$ & $\underline{0.229 \pm 0.103}$
& $\underline{31.87 \pm 2.08}$ & $\underline{0.871 \pm 0.033}$ & $\underline{0.110 \pm 0.032}$ \\

& RED-diff \cite{mardani2023variational} 
& $23.42 \pm 2.68$ & $0.589 \pm 0.128$ & $0.438 \pm 0.125$
& $26.79 \pm 1.66$ & $0.755 \pm 0.053$ & $0.320 \pm 0.053$ \\

& DAPS \cite{zhang2025improving}
& $25.06 \pm 3.00$ & $0.609 \pm 0.119$ & $0.326 \pm 0.072$
& $28.09 \pm 1.90$ & $0.734 \pm 0.040$ & $0.217 \pm 0.032$ \\

\cmidrule(lr){2-8}

& OptDiff (Ours)
& $26.27 \pm 3.24$ & $0.716 \pm 0.110$ & $0.243 \pm 0.085$
& $30.77 \pm 1.75$ & $0.848 \pm 0.031$ & $0.111 \pm 0.029$ \\

\midrule

\multirow{5}{*}{Super-res $4\times$}

& DDNM \cite{wang2022zero}
& $25.23 \pm 3.64$ & $0.681 \pm 0.154$ & $0.286 \pm 0.114$
& $30.00 \pm 2.27$ & $0.851 \pm 0.044$ & $\mathbf{0.110 \pm 0.028}$ \\

& DiffPIR \cite{zhu2023denoising}
& $25.38 \pm 3.65$ & $0.685 \pm 0.144$ & $\underline{0.281 \pm 0.127}$
& $\underline{31.05 \pm 2.38}$ & $\underline{0.872 \pm 0.039}$ & $\underline{0.112 \pm 0.033}$ \\

& RED-diff \cite{mardani2023variational}
& $24.74 \pm 3.31$ & $0.672 \pm 0.147$ & $0.307 \pm 0.108$
& $29.54 \pm 2.08$ & $0.831 \pm 0.041$ & $0.144 \pm 0.031$ \\

& DAPS \cite{zhang2025improving}
& $\mathbf{26.14 \pm 3.24}$ & $\underline{0.697 \pm 0.116}$ & $\mathbf{0.229 \pm 0.055}$
& $29.86 \pm 2.17$ & $0.831 \pm 0.038$ & $0.132 \pm 0.023$ \\

\cmidrule(lr){2-8}

& OptDiff (Ours)
& $\underline{25.82 \pm 3.80}$ & $\mathbf{0.705 \pm 0.145}$ & $0.340 \pm 0.146$
& $\mathbf{31.23 \pm 2.57}$ & $\mathbf{0.879 \pm 0.039}$ & $0.148 \pm 0.035$ \\

\midrule

\multirow{5}{*}{Super-res $16\times$}

& DDNM \cite{wang2022zero}
& $19.38 \pm 2.52$ & $0.422 \pm 0.172$ & $\underline{0.560 \pm 0.159}$
& $22.25 \pm 2.06$ & $\underline{0.637 \pm 0.084}$ & $\mathbf{0.258 \pm 0.069}$ \\

& DiffPIR \cite{zhu2023denoising}
& $19.81 \pm 2.58$ & $\underline{0.444 \pm 0.176}$ & $0.648 \pm 0.181$
& $\underline{22.30 \pm 2.01}$ & $\mathbf{0.640 \pm 0.082}$ & $\underline{0.269 \pm 0.073}$ \\

& RED-diff \cite{mardani2023variational}
& $19.83 \pm 2.53$ & $0.439 \pm 0.169$ & $0.680 \pm 0.145$
& $\mathbf{22.57 \pm 2.08}$ & $0.634 \pm 0.079$ & $0.348 \pm 0.079$ \\

& DAPS \cite{zhang2025improving}
& $\mathbf{20.23 \pm 2.43}$ & $0.440 \pm 0.150$ & $\mathbf{0.521 \pm 0.078}$
& $22.24 \pm 1.98$ & $0.598 \pm 0.072$ & $0.342 \pm 0.047$ \\

\cmidrule(lr){2-8}

& OptDiff (Ours)
& $\underline{20.19 \pm 2.70}$ & $\mathbf{0.453 \pm 0.177}$ & $0.715 \pm 0.173$
& $\mathbf{22.57 \pm 2.16}$ & $0.634 \pm 0.083$ & $0.421 \pm 0.082$ \\

\midrule

\multirow{5}{*}{\shortstack{Inpainting \\ $64 \times 64$}}

& DDNM \cite{wang2022zero}
& $26.28 \pm 3.03$ & $0.910 \pm 0.020$ & $\underline{0.055 \pm 0.016}$
& $\underline{33.26 \pm 2.63}$ & $0.961 \pm 0.009$ & $0.021 \pm 0.006$ \\

& DiffPIR \cite{zhu2023denoising}
& $\mathbf{28.35 \pm 3.63}$ & $\mathbf{0.935 \pm 0.012}$ & $\mathbf{0.049 \pm 0.018}$
& $\mathbf{34.65 \pm 2.85}$ & $\mathbf{0.972 \pm 0.007}$ & $\mathbf{0.010 \pm 0.004}$ \\

& RED-diff \cite{mardani2023variational}
& $24.45 \pm 2.52$ & $0.919 \pm 0.011$ & $0.080 \pm 0.015$
& $26.32 \pm 2.14$ & $0.947 \pm 0.007$ & $0.066 \pm 0.010$ \\

& DAPS \cite{zhang2025improving}
& $\underline{27.17 \pm 3.25}$ & $0.922 \pm 0.013$ & $0.066 \pm 0.031$
& $33.21 \pm 2.70$ & $0.964 \pm 0.007$ & $0.023 \pm 0.009$ \\

\cmidrule(lr){2-8}

& OptDiff (Ours)
& $26.61 \pm 3.02$ & $\underline{0.926 \pm 0.013}$ & $0.064 \pm 0.015$
& $33.12 \pm 3.39$ & $\underline{0.968 \pm 0.008}$ & $\underline{0.017 \pm 0.007}$ \\

\bottomrule
\end{tabular}
}
\end{table}

\begin{table}[t]
\centering
\caption{
Comparison with baselines using NFE-100. Our method consistently achieves the best or comparable performance across all task--dataset pairs. Best results are shown in bold, and second-best results are underlined.}
\label{tab:results-100steps}
\resizebox{\linewidth}{!}{%
\begin{tabular}{c l ccc ccc}

\toprule
& & \multicolumn{6}{c}{\textbf{fastMRI}} \\
& & \multicolumn{3}{c}{\textbf{Acceleration R=4}} & \multicolumn{3}{c}{\textbf{Acceleration R=8}} \\
\cmidrule(lr){3-5} \cmidrule(lr){6-8}
\textbf{Task} & \textbf{Method} 
& \textbf{PSNR} $\uparrow$ & \textbf{SSIM} $\uparrow$ & \cellcolor{gray!20}{}
& \textbf{PSNR} $\uparrow$ & \textbf{SSIM} $\uparrow$ & \cellcolor{gray!20}{} \\
\midrule

\multirow{7}{*}{\shortstack{MRI recon}}

& DPS \cite{chung2022diffusion}
& $35.82 \pm 1.46$ & $0.775 \pm 0.069$ & \cellcolor{gray!20}{}
& $28.94 \pm 1.54$ & $0.774 \pm 0.066$ & \cellcolor{gray!20}{} \\

& DiffPIR \cite{zhu2023denoising}
& $30.33 \pm 0.97$&  ${0.691 \pm 0.077}$& \cellcolor{gray!20}{}
& $27.17 \pm 1.08$ & $0.650 \pm 0.072$ & \cellcolor{gray!20}{} \\

& Pi-GDM \cite{song2023pseudoinverse}
& $30.69 \pm 1.37$ & $0.753 \pm 0.086$ & \cellcolor{gray!20}{}
& $28.88 \pm 1.27$ & $0.734 \pm 0.078$ & \cellcolor{gray!20}{} \\

& RED-diff \cite{mardani2023variational}
& $\underline{36.79 \pm 1.43}$ & $0.872 \pm 0.035$ & \cellcolor{gray!20}{}
& $\underline{31.92 \pm 1.44}$ & $0.840 \pm 0.034$ & \cellcolor{gray!20}{} \\

& DDS \cite{chung2024decomposed}
& $33.72 \pm 1.14$ & $\underline{0.930 \pm 0.030}$ & \cellcolor{gray!20}{}
& $28.92 \pm 1.28$ & $0.828 \pm 0.027$ & \cellcolor{gray!20}{} \\

& DAPS \cite{zhang2025improving}
& $36.50 \pm 1.63$ & $0.915 \pm 0.033$ & \cellcolor{gray!20}{}
& $31.66 \pm 1.49$ & $\underline{0.870 \pm 0.042}$ & \cellcolor{gray!20}{} \\

\cmidrule(lr){2-8}

& OptDiff (Ours)
& $\mathbf{38.45 \pm 1.58}$ & $\mathbf{0.948 \pm 0.023}$ & \cellcolor{gray!20}{}
& $\mathbf{34.93 \pm 1.38}$ & $\mathbf{0.921 \pm 0.029}$ & \cellcolor{gray!20}{} \\
\bottomrule

\toprule
& & \multicolumn{3}{c}{\textbf{ImageNet}} & \multicolumn{3}{c}{\textbf{FFHQ}} \\
\cmidrule(lr){3-5} \cmidrule(lr){6-8}
\textbf{Task} & \textbf{Method} 
& \textbf{PSNR} $\uparrow$ & \textbf{SSIM} $\uparrow$ & \textbf{LPIPS} $\downarrow$
& \textbf{PSNR} $\uparrow$ & \textbf{SSIM} $\uparrow$ & \textbf{LPIPS} $\downarrow$ \\
\cmidrule(lr){1-8}

\multirow{9}{*}{\shortstack{Deblurring \\ kernel $6$}}

& DPS \cite{chung2022diffusion}
& $24.25 \pm 3.97$ & $0.602 \pm 0.192$ & $0.420 \pm 0.187$
& $33.00 \pm 2.50$ & $0.898 \pm 0.034$ & $0.078 \pm 0.028$ \\

& SITCOM \cite{alkhouri2025sitcom}
& $28.60 \pm 3.27$ & $0.792 \pm 0.060$ & $0.099 \pm 0.049$
& $32.41 \pm 2.17$ & $0.878 \pm 0.031$ & $0.081 \pm 0.024$ \\

& DMPlug \cite{wang2024dmplug}
& $25.83 \pm 3.32$ & $0.684 \pm 0.141$ & $0.268 \pm 0.104$
& $30.19 \pm 1.94$ & $0.840 \pm 0.046$ & $0.177 \pm 0.044$ \\

& DDRM \cite{kawar2022denoising}    
& $\underline{32.71 \pm 3.82}$ & $\underline{0.908 \pm 0.045}$ & $\underline{0.039 \pm 0.017}$
& $\underline{36.96 \pm 1.76}$ & $\mathbf{0.949 \pm 0.013}$ & $\underline{0.030 \pm 0.009}$ \\

& DDNM \cite{wang2022zero}          
& $32.60 \pm 3.89$ & $0.906 \pm 0.045$ & $\mathbf{0.030 \pm 0.013}$
& $36.95 \pm 1.78$ & $\mathbf{0.949 \pm 0.013}$ & $\mathbf{0.027 \pm 0.008}$ \\

& DiffPIR \cite{zhu2023denoising}   
& $32.22 \pm 3.68$ & $0.892 \pm 0.049$ & $0.061 \pm 0.032$
& $36.58 \pm 1.76$ & $\underline{0.942 \pm 0.014}$ & $0.040 \pm 0.012$ \\

& RED-diff \cite{mardani2023variational} 
& $29.35 \pm 3.92$ & $0.821 \pm 0.092$ & $0.175 \pm 0.089$
& $35.07 \pm 1.77$ & $0.915 \pm 0.015$ & $0.063 \pm 0.020$ \\

& DAPS \cite{zhang2025improving}    
& $31.39 \pm 3.13$ & $0.853 \pm 0.047$ & $0.107 \pm 0.033$
& $34.13 \pm 1.39$ & $0.888 \pm 0.013$ & $0.088 \pm 0.021$ \\

\cmidrule(lr){2-8}

& OptDiff (Ours)     
& $\mathbf{33.28 \pm 3.74}$ & $\mathbf{0.914 \pm 0.038}$ & $0.049 \pm 0.024$
& $\mathbf{37.10 \pm 1.79}$ & $\mathbf{0.949 \pm 0.013}$ & $0.058 \pm 0.017$ \\

\midrule

\multirow{9}{*}{\shortstack{Severe deblur \\ kernel $12$}}

& DPS \cite{chung2022diffusion}
& $21.30 \pm 3.27$ & $0.484 \pm 0.200$ & $0.371 \pm 0.159$
& $25.88 \pm 6.47$ & $0.749 \pm 0.129$ & $0.159 \pm 0.115$ \\

& SITCOM \cite{alkhouri2025sitcom}
& $24.25 \pm 3.27$ & $0.596 \pm 0.142$ & $0.285 \pm 0.118$
& $28.65 \pm 2.29$ & $0.797 \pm 0.054$ & $0.157 \pm 0.046$ \\

& DMPlug \cite{wang2024dmplug}
& $22.72 \pm 3.01$ & $0.547 \pm 0.176$ & $0.390 \pm 0.141$
& $27.02 \pm 2.21$ & $0.763 \pm 0.070$ & $0.235 \pm 0.058$ \\

& DDRM \cite{kawar2022denoising}    
& $27.87 \pm 3.89$ & $0.785 \pm 0.110$ & $0.156 \pm 0.074$
& $32.18 \pm 2.18$ & $0.889 \pm 0.031$ & $\underline{0.069 \pm 0.021}$ \\

& DDNM \cite{wang2022zero}
& $\underline{28.07 \pm 4.12}$ & $\mathbf{0.789 \pm 0.110}$ & $\mathbf{0.134 \pm 0.059}$
& $\underline{32.54 \pm 2.15}$ & $\mathbf{0.893 \pm 0.030}$ & $0.074 \pm 0.022$ \\

& DiffPIR \cite{zhu2023denoising}
& $27.41 \pm 3.72$ & $0.746 \pm 0.114$ & $\underline{0.145 \pm 0.064}$
& $32.05 \pm 1.95$ & $0.870 \pm 0.030$ & $\mathbf{0.060 \pm 0.019}$ \\

& RED-diff  \cite{mardani2023variational}
& $24.19 \pm 3.49$ & $0.604 \pm 0.171$ & $0.428 \pm 0.156$
& $29.42 \pm 2.23$ & $0.800 \pm 0.048$ & $0.183 \pm 0.054$ \\

& DAPS  \cite{zhang2025improving}    
& $26.28 \pm 2.80$ & $0.671 \pm 0.086$ & $0.247 \pm 0.075$
& $28.38 \pm 2.19$ & $0.786 \pm 0.055$ & $0.199 \pm 0.040$ \\

\cmidrule(lr){2-8}

& OptDiff (Ours) 
& $\mathbf{28.27 \pm 3.90}$ & $\underline{0.788 \pm 0.100}$ & $0.190 \pm 0.083$
& $\mathbf{32.61 \pm 2.22}$ & $\underline{0.891 \pm 0.031}$ & $0.127 \pm 0.034$ \\

\midrule

\multirow{9}{*}{Super-res $4\times$}

& DPS \cite{chung2022diffusion}
& $23.72 \pm 4.90$ & $0.618 \pm 0.209$ & $0.299 \pm 0.207$
& $30.50 \pm 2.62$ & $0.866 \pm 0.044$ & $\mathbf{0.071 \pm 0.025}$ \\

& SITCOM \cite{alkhouri2025sitcom}
& $25.75 \pm 3.91$ & $0.696 \pm 0.146$ & $0.286 \pm 0.141$
& $31.04 \pm 2.49$ & $0.863 \pm 0.040$ & $0.091 \pm 0.029$ \\

& DMPlug \cite{wang2024dmplug}
& $25.03 \pm 3.78$ & $0.665 \pm 0.160$ & $\mathbf{0.224 \pm 0.084}$
& $30.36 \pm 2.23$ & $0.854 \pm 0.042$ & $0.151 \pm 0.036$ \\

& DDRM \cite{kawar2022denoising}     
& $25.84 \pm 3.86$ & $0.702 \pm 0.152$ & $0.331 \pm 0.145$
& $\underline{31.68 \pm 2.50}$ & $\underline{0.885 \pm 0.038}$ & $0.125 \pm 0.038$ \\

& DDNM \cite{wang2022zero}           
& $25.79 \pm 3.93$ & $0.703 \pm 0.152$ & $0.286 \pm 0.136$
& $30.94 \pm 2.45$ & $0.872 \pm 0.041$ & $\underline{0.085 \pm 0.028}$ \\

& DiffPIR \cite{zhu2023denoising}    
& $25.28 \pm 3.78$ & $0.682 \pm 0.152$ & $0.242 \pm 0.117$
& $31.06 \pm 2.44$ & $0.873 \pm 0.040$ & $0.092 \pm 0.031$ \\

& RED-diff  \cite{mardani2023variational} 
& $25.75 \pm 3.82$ & $0.700 \pm 0.141$ & $0.311 \pm 0.144$
& ${31.25 \pm 2.37}$ & $0.875 \pm 0.037$ & $0.108 \pm 0.034$ \\

& DAPS \cite{zhang2025improving}    
& $\mathbf{26.50 \pm 3.43}$ & $\underline{0.706 \pm 0.132}$ & $\underline{0.239 \pm 0.067}$
& $29.63 \pm 1.73$ & $0.763 \pm 0.028$ & $0.179 \pm 0.028$ \\

\cmidrule(lr){2-8}

& OptDiff (Ours)     
& $\underline{26.07 \pm 3.96}$ & $\mathbf{0.715 \pm 0.148}$ & $0.350 \pm 0.156$
& $\mathbf{31.99 \pm 2.57}$ & $\mathbf{0.891 \pm 0.037}$ & $0.147 \pm 0.043$ \\

\midrule

\multirow{9}{*}{Super-res $16\times$}

& DPS \cite{chung2022diffusion}
& $17.95 \pm 3.02$ & $0.365 \pm 0.192$ & $\mathbf{0.388 \pm 0.125}$
& $22.62 \pm 2.28$ & $0.664 \pm 0.086$ & $0.286 \pm 0.078$ \\

& SITCOM \cite{alkhouri2025sitcom}
& $20.25 \pm 2.81$ & $\underline{0.463 \pm 0.185}$ & $0.616 \pm 0.191$
& $23.49 \pm 2.36$ & $\mathbf{0.689 \pm 0.085}$ & $0.263 \pm 0.073$ \\

& DMPlug \cite{wang2024dmplug}
& $19.57 \pm 2.63$ & $0.431 \pm 0.177$ & $\underline{0.517 \pm 0.142}$
& $22.75 \pm 2.47$ & $0.646 \pm 0.085$ & $0.267 \pm 0.057$ \\

& DDRM \cite{kawar2022denoising}     
& $20.35 \pm 2.77$ & ${0.462 \pm 0.185}$ & $0.683 \pm 0.199$
& $\underline{23.50 \pm 2.24}$ & $\underline{0.687 \pm 0.083}$ & $0.316 \pm 0.090$ \\

& DDNM \cite{wang2022zero}
& $20.15 \pm 2.68$ & $0.460 \pm 0.183$ & $0.633 \pm 0.207$
& $22.86 \pm 2.12$ & $0.663 \pm 0.083$ & $\underline{0.249 \pm 0.079}$ \\

& DiffPIR \cite{zhu2023denoising}
& $19.66 \pm 2.61$ & $0.439 \pm 0.178$ & $0.577 \pm 0.194$
& $22.77 \pm 2.14$ & $0.658 \pm 0.085$ & $\mathbf{0.246 \pm 0.077}$ \\

& RED-diff  \cite{mardani2023variational}
& $20.34 \pm 2.76$ & $0.460 \pm 0.182$ & $0.704 \pm 0.193$
& $23.01 \pm 2.07$ & $0.663 \pm 0.082$ & $0.282 \pm 0.082$ \\

& DAPS  \cite{zhang2025improving}     
& $\mathbf{20.53 \pm 2.49}$ & $0.418 \pm 0.136$ & ${0.522 \pm 0.081}$
& $22.63 \pm 1.96$ & $0.572 \pm 0.064$ & $0.358 \pm 0.046$ \\

\cmidrule(lr){2-8}

& OptDiff (Ours) 
& $\underline{20.44 \pm 2.75}$ & $\mathbf{0.468 \pm 0.184}$ & $0.704 \pm 0.186$
& $\mathbf{23.67 \pm 2.25}$ & $\mathbf{0.689 \pm 0.083}$ & $0.343 \pm 0.094$ \\
\midrule

\multirow{9}{*}{\shortstack{Inpainting \\ $64 \times 64$}}

& DPS \cite{chung2022diffusion}
& $23.41 \pm 3.49$ & $0.649 \pm 0.186$ & $0.349 \pm 0.176$
& $32.09 \pm 3.84$ & $0.931 \pm 0.030$ & $0.074 \pm 0.037$ \\

& SITCOM \cite{alkhouri2025sitcom}
& $28.87 \pm 3.65$ & $0.929 \pm 0.015$ & $\underline{0.038 \pm 0.015}$
& $\mathbf{36.11 \pm 2.33}$ & ${0.973 \pm 0.006}$ & $\mathbf{0.009 \pm 0.004}$ \\

& DMPlug \cite{wang2024dmplug}
& $21.41 \pm 3.19$ & $0.765 \pm 0.091$ & $0.246 \pm 0.098$
& $29.64 \pm 3.10$ & $0.881 \pm 0.029$ & $0.143 \pm 0.039$ \\

& DDRM \cite{kawar2022denoising}
& $\underline{29.02 \pm 3.68}$ & $\underline{0.935 \pm 0.016}$ & $0.053 \pm 0.018$
& $\underline{35.42 \pm 2.70}$ & $\mathbf{0.978 \pm 0.006}$ & $\mathbf{0.009 \pm 0.004}$ \\
& DDNM \cite{wang2022zero}
& $28.07 \pm 3.61$ & $0.928 \pm 0.013$ & $\mathbf{0.036 \pm 0.014}$
& ${34.53 \pm 2.57}$ & $0.970 \pm 0.007$ & $\underline{0.010 \pm 0.004}$ \\

& DiffPIR \cite{zhu2023denoising}
& $\mathbf{29.18 \pm 3.89}$ & $\mathbf{0.944 \pm 0.012}$ & $0.047 \pm 0.019$
& $35.40 \pm 2.62$ & $\underline{0.975 \pm 0.006}$ & $\underline{0.010 \pm 0.004}$ \\

& RED-diff  \cite{mardani2023variational}
& $27.70 \pm 3.02$ & $0.932 \pm 0.010$ & $0.058 \pm 0.016$
& $29.49 \pm 1.51$ & $0.931 \pm 0.009$ & $0.035 \pm 0.008$ \\

& DAPS  \cite{zhang2025improving}     
& $28.80 \pm 3.64$ & $0.931 \pm 0.011$ & $0.059 \pm 0.022$
& ${34.53 \pm 2.79}$ & $0.967 \pm 0.007$ & ${0.021 \pm 0.009}$ \\

\cmidrule(lr){2-8}

& OptDiff (Ours) 
& $28.36 \pm 3.38$ & $0.930 \pm 0.014$ & $0.054 \pm 0.018$
& $33.40 \pm 3.77$ & $0.966 \pm 0.019$ & $0.023 \pm 0.023$ \\

\bottomrule
\end{tabular}
}
\end{table}

\clearpage
\section{Runtime comparisons}

\begin{table}[h]
\centering
\caption{
Runtime per reconstruction in seconds at NFE-20 and NFE-100. Mean and standard deviation are calculated on a single NVIDIA A40 GPU with batch size 1, over 20 reconstructions.}
\label{tab:runtime-combined}
\resizebox{0.9\linewidth}{!}{%
\begin{tabular}{c l cc cc}

\toprule
& & \multicolumn{4}{c}{\textbf{fastMRI}} \\
& & \multicolumn{2}{c}{\textbf{Acceleration R=4}} & \multicolumn{2}{c}{\textbf{Acceleration R=8}} \\
\cmidrule(lr){3-4} \cmidrule(lr){5-6}
\textbf{Task} & \textbf{Method}
& \textbf{NFE-20} & \textbf{NFE-100}
& \textbf{NFE-20} & \textbf{NFE-100} \\
\midrule

\multirow{7}{*}{\shortstack{MRI recon}}

& DPS \cite{chung2022diffusion}
& $3.30 \pm 0.00$ & $16.57 \pm 0.01$
& $3.30 \pm 0.00$ & $16.55 \pm 0.01$ \\

& DiffPIR \cite{zhu2023denoising}
& $1.63 \pm 0.00$ & $8.14 \pm 0.01$
& $1.63 \pm 0.00$ & $8.15 \pm 0.01$ \\

& Pi-GDM \cite{song2023pseudoinverse}
& $3.81 \pm 0.00$ & $19.05 \pm 0.01$
& $3.81 \pm 0.00$ & $19.06 \pm 0.01$ \\

& RED-diff \cite{mardani2023variational}
& $1.63 \pm 0.00$ & $8.12 \pm 0.01$
& $1.63 \pm 0.00$ & $8.12 \pm 0.01$ \\

& DAPS \cite{zhang2025improving}
& $1.79 \pm 0.01$ & $8.38 \pm 0.01$
& $1.79 \pm 0.00$ & $8.39 \pm 0.02$ \\

\cmidrule(lr){2-6}

& OptDiff (Ours)
& $1.67 \pm 0.01$ & $8.36 \pm 0.02$
& $1.67 \pm 0.01$ & $8.36 \pm 0.01$ \\

\bottomrule

\toprule
& & \multicolumn{2}{c}{\textbf{ImageNet}} & \multicolumn{2}{c}{\textbf{FFHQ}} \\
\cmidrule(lr){3-4} \cmidrule(lr){5-6}
\textbf{Task} & \textbf{Method}
& \textbf{NFE-20} & \textbf{NFE-100}
& \textbf{NFE-20} & \textbf{NFE-100} \\
\cmidrule(lr){1-6}

\multirow{7}{*}{\shortstack{Deblurring \\ kernel $6$}}

& DPS \cite{chung2022diffusion}
& $5.33 \pm 0.11$ & $26.29 \pm 0.03$
& $1.68 \pm 0.00$ & $8.39 \pm 0.04$ \\

& DDRM \cite{kawar2022denoising}
& $1.48 \pm 0.00$ & $7.56 \pm 0.25$
& $0.63 \pm 0.00$ & $3.15 \pm 0.01$ \\

& DDNM \cite{wang2022zero}
& $1.48 \pm 0.00$ & $7.43 \pm 0.01$
& $0.64 \pm 0.00$ & $3.14 \pm 0.06$ \\

& DiffPIR \cite{zhu2023denoising}
& $1.46 \pm 0.00$ & $7.34 \pm 0.01$
& $0.62 \pm 0.00$ & $3.07 \pm 0.03$ \\

& RED-diff \cite{mardani2023variational}
& $1.46 \pm 0.00$ & $7.34 \pm 0.01$
& $0.65 \pm 0.00$ & $3.12 \pm 0.02$ \\

& DAPS \cite{zhang2025improving}
& $5.36 \pm 0.05$ & $32.59 \pm 0.73$
& $1.27 \pm 0.09$ & $28.25 \pm 0.38$ \\
\cmidrule(lr){2-6}

& OptDiff (Ours)
& $1.48 \pm 0.00$ & $7.33 \pm 0.01$
& $0.64 \pm 0.03$ & $3.15 \pm 0.04$ \\

\midrule

\multirow{7}{*}{\shortstack{Severe deblur \\ kernel $12$}}

& DPS \cite{chung2022diffusion}
& $5.26 \pm 0.01$ & $26.39 \pm 0.07$
& $1.68 \pm 0.02$ & $8.40 \pm 0.02$ \\

& DDRM \cite{kawar2022denoising}
& $1.48 \pm 0.00$ & $7.42 \pm 0.01$
& $0.63 \pm 0.00$ & $3.22 \pm 0.01$ \\

& DDNM \cite{wang2022zero}
& $1.48 \pm 0.00$ & $7.44 \pm 0.01$
& $0.65 \pm 0.04$ & $3.14 \pm 0.06$ \\

& DiffPIR \cite{zhu2023denoising}
& $1.46 \pm 0.00$ & $7.36 \pm 0.02$
& $0.63 \pm 0.04$ & $3.14 \pm 0.01$ \\

& RED-diff \cite{mardani2023variational}
& $1.47 \pm 0.02$ & $7.34 \pm 0.01$
& $0.65 \pm 0.00$ & $3.13 \pm 0.03$ \\

& DAPS \cite{zhang2025improving}
& $5.44 \pm 0.08$ & $24.52 \pm 0.75$
& $4.53 \pm 0.03$ & $28.08 \pm 0.51$ \\

\cmidrule(lr){2-6}

& OptDiff (Ours)
& $1.48 \pm 0.01$ & $7.43 \pm 0.04$
& $0.64 \pm 0.03$ & $3.15 \pm 0.01$ \\

\midrule

\multirow{7}{*}{Super-res $4\times$}

& DPS \cite{chung2022diffusion}
& $5.21 \pm 0.01$ & $26.29 \pm 0.02$
& $1.66 \pm 0.00$ & $8.35 \pm 0.03$ \\

& DDRM \cite{kawar2022denoising}
& $1.48 \pm 0.00$ & $7.41 \pm 0.01$
& $0.63 \pm 0.00$ & $3.20 \pm 0.02$ \\

& DDNM \cite{wang2022zero}
& $1.49 \pm 0.03$ & $7.42 \pm 0.01$
& $0.65 \pm 0.00$ & $3.13 \pm 0.02$ \\

& DiffPIR \cite{zhu2023denoising}
& $1.46 \pm 0.00$ & $7.34 \pm 0.01$
& $0.61 \pm 0.00$ & $3.12 \pm 0.02$ \\

& RED-diff \cite{mardani2023variational}
& $1.46 \pm 0.00$ & $7.32 \pm 0.01$
& $0.64 \pm 0.02$ & $3.09 \pm 0.00$ \\

& DAPS \cite{zhang2025improving}
& $1.77 \pm 0.02$ & $22.82 \pm 0.60$
& $1.13 \pm 0.03$ & $18.28 \pm 0.71$ \\

\cmidrule(lr){2-6}

& OptDiff (Ours)
& $1.47 \pm 0.00$ & $7.31 \pm 0.02$
& $0.64 \pm 0.05$ & $3.06 \pm 0.07$ \\

\midrule

\multirow{7}{*}{Super-res $16\times$}

& DPS \cite{chung2022diffusion}
& $5.21 \pm 0.01$ & $26.21 \pm 0.02$
& $1.68 \pm 0.00$ & $8.33 \pm 0.02$ \\

& DDRM \cite{kawar2022denoising}
& $1.47 \pm 0.00$ & $7.42 \pm 0.02$
& $0.63 \pm 0.00$ & $3.11 \pm 0.03$ \\

& DDNM \cite{wang2022zero}
& $1.48 \pm 0.02$ & $7.42 \pm 0.01$
& $0.65 \pm 0.00$ & $3.17 \pm 0.10$ \\

& DiffPIR \cite{zhu2023denoising}
& $1.47 \pm 0.01$ & $7.35 \pm 0.01$
& $0.62 \pm 0.01$ & $3.04 \pm 0.05$ \\

& RED-diff \cite{mardani2023variational}
& $1.46 \pm 0.01$ & $7.32 \pm 0.01$
& $0.63 \pm 0.00$ & $3.11 \pm 0.03$ \\

& DAPS \cite{zhang2025improving}
& $1.80 \pm 0.08$ & $22.37 \pm 1.44$
& $0.97 \pm 0.05$ & $18.37 \pm 0.23$ \\

\cmidrule(lr){2-6}

& OptDiff (Ours)
& $1.47 \pm 0.02$ & $7.33 \pm 0.01$
& $0.62 \pm 0.03$ & $3.08 \pm 0.01$ \\

\midrule

\multirow{7}{*}{\shortstack{Inpainting \\ $64 \times 64$}}

& DPS \cite{chung2022diffusion}
& $5.27 \pm 0.03$ & $26.32 \pm 0.06$
& $1.67 \pm 0.00$ & $8.38 \pm 0.02$ \\

& DDRM \cite{kawar2022denoising}
& $1.47 \pm 0.00$ & $7.40 \pm 0.01$
& $0.62 \pm 0.01$ & $3.09 \pm 0.02$ \\

& DDNM \cite{wang2022zero}
& $1.47 \pm 0.00$ & $7.37 \pm 0.04$
& $0.63 \pm 0.00$ & $3.05 \pm 0.02$ \\

& DiffPIR \cite{zhu2023denoising}
& $1.46 \pm 0.01$ & $7.35 \pm 0.01$
& $0.61 \pm 0.00$ & $3.05 \pm 0.04$ \\

& RED-diff \cite{mardani2023variational}
& $1.46 \pm 0.01$ & $7.33 \pm 0.03$
& $0.66 \pm 0.05$ & $3.17 \pm 0.02$ \\

& DAPS \cite{zhang2025improving}
& $1.68 \pm 0.06$ & $13.80 \pm 1.30$
& $0.68 \pm 0.00$ & $9.81 \pm 0.73$ \\

\cmidrule(lr){2-6}

& OptDiff (Ours)
& $1.46 \pm 0.00$ & $7.33 \pm 0.01$
& $0.62 \pm 0.02$ & $3.06 \pm 0.01$ \\

\bottomrule
\end{tabular}%
}
\end{table}

\clearpage
\section{Additional Qualitative Results}
\label{additional_results}
We provide visual comparisons on additional test set slices for the MRI reconstruction, super-resolution, deblurring, inpainting tasks.
\begin{figure}[h]
  \centering
  \includegraphics[width=15cm]{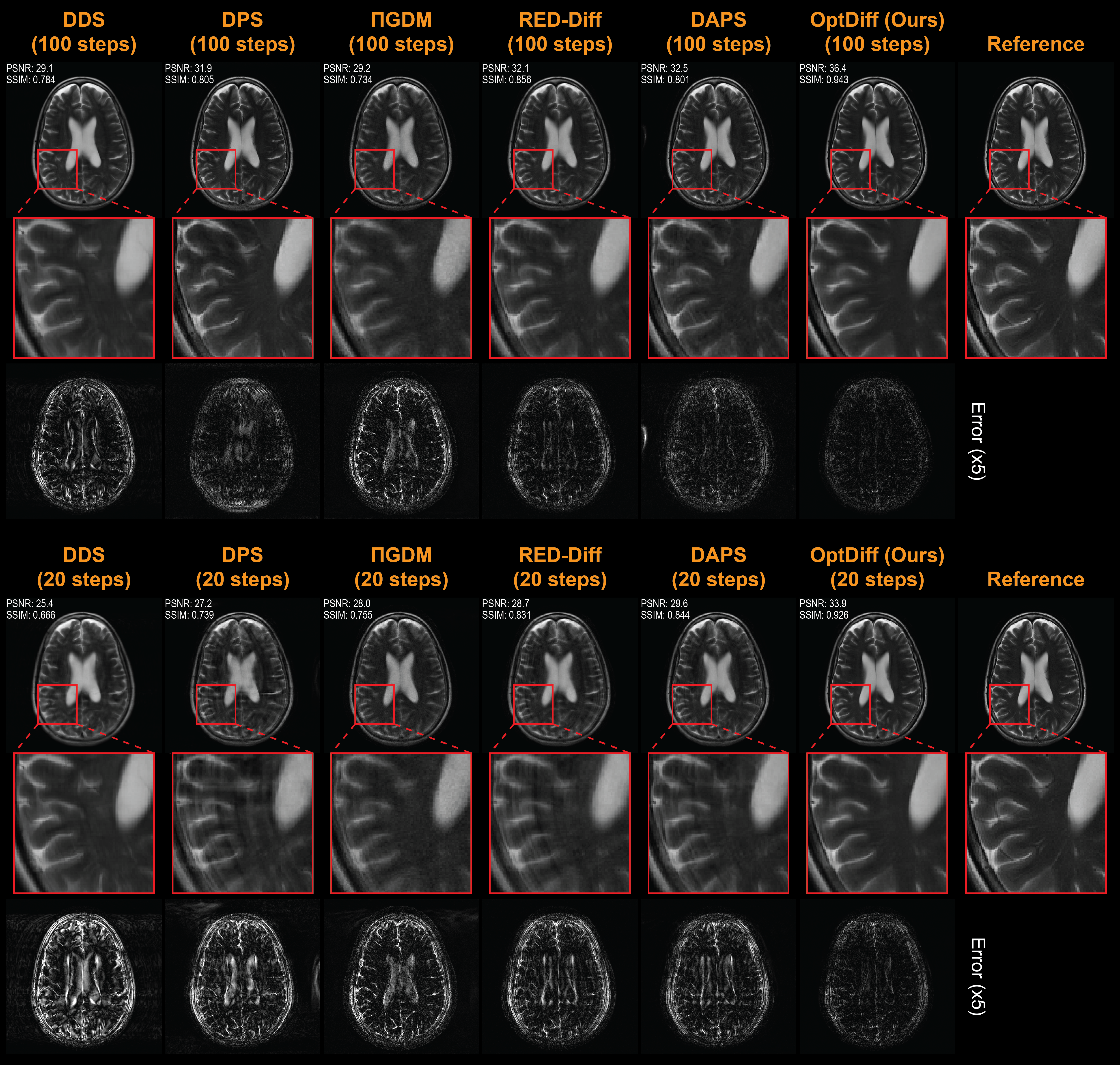}
  \caption{Reconstruction performance comparison on a representative test slice for the MRI Reconstruction task with $R=8$ random subsampling on the fastMRI dataset.} \label{fig:comparisons_mri_appdx}
\end{figure}
\begin{figure}[h]
  \centering
  \includegraphics[width=16cm]{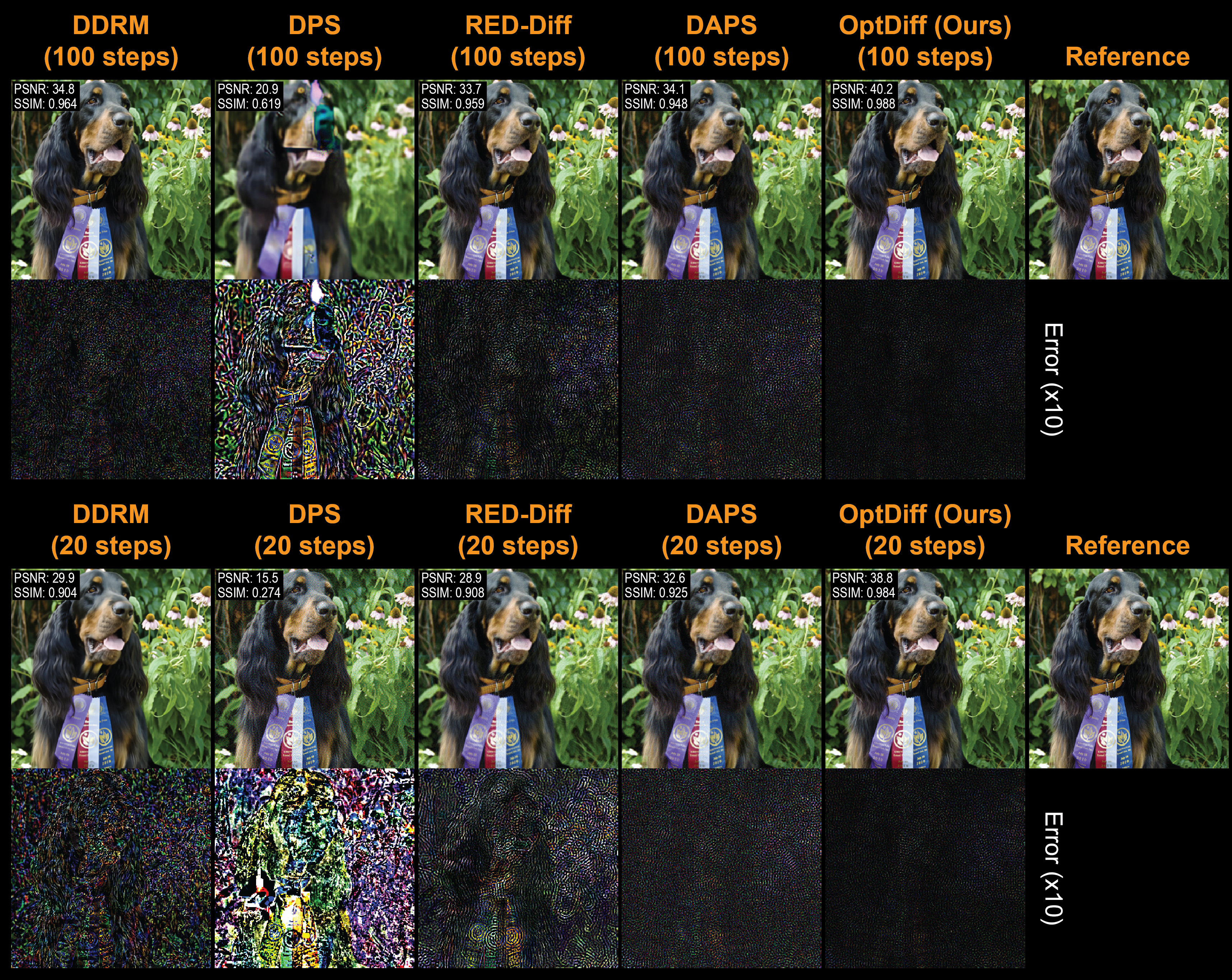}
  \caption{Reconstruction performance comparison on a representative test slice for deblurring task on the ImageNet dataset.} \label{fig:comparisons_deblurring_appdx}
\end{figure}
\begin{figure}[h]
  \centering
  \includegraphics[width=16cm]{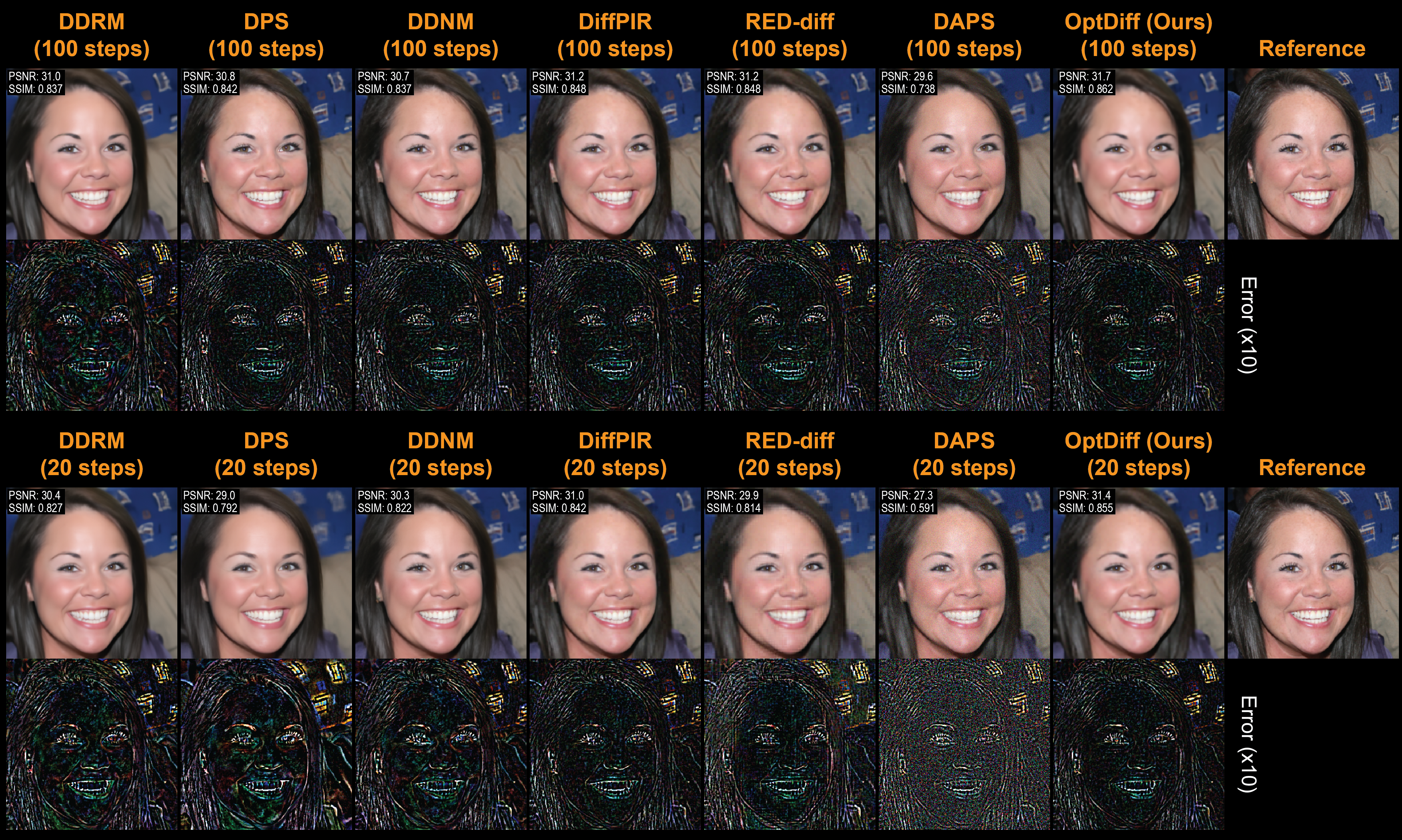}
  \caption{Reconstruction performance comparison on a representative test slice for $4\times$ super-resolution task on the FFHQ dataset.} \label{fig:comparisons_superres_appdx}
\end{figure}
\begin{figure}[h]
  \centering
  \includegraphics[width=16cm]{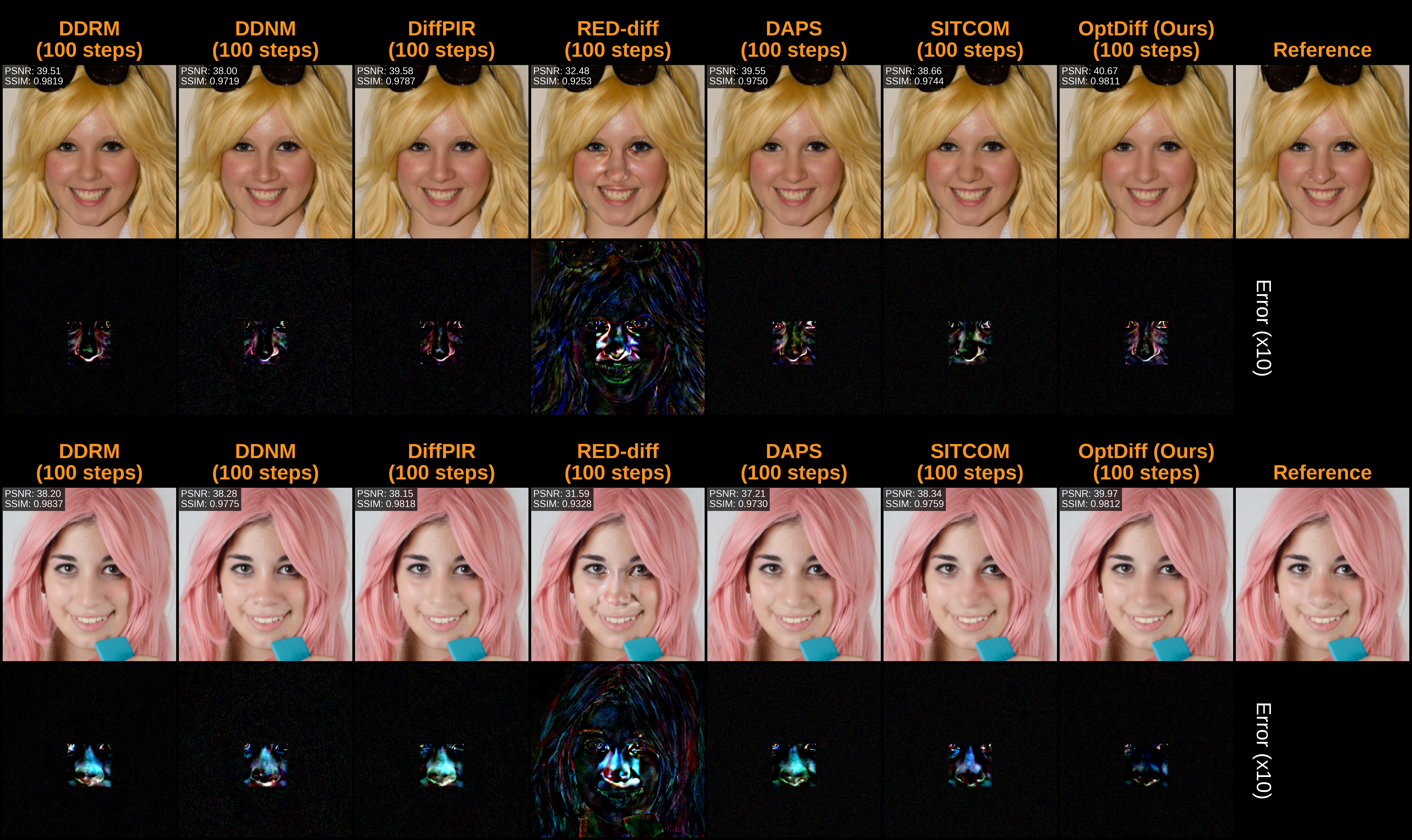}
  \caption{Reconstruction performance comparison on representative test slices for inpainting task on the FFHQ dataset.} \label{fig:comparisons_inpainting_appdx}
\end{figure}
%

\par
\clearpage

\section{Broader Impacts}

The methods developed in this work have the potential to significantly improve MRI image reconstruction quality and efficiency, which could translate into meaningful clinical benefits. Improved reconstructions may lead to more accurate diagnoses and better patient outcomes. Additionally, by enabling faster acquisitions, these methods could enhance scanner throughput and increase access to MRI, particularly in resource-limited settings.

However, there are also potential risks and limitations to consider. This study relies on the publicly available fastMRI dataset, and while it has become a benchmark in the community, it may not fully capture the diversity of patient populations, scanner hardware, or clinical conditions. This raises concerns about potential bias and generalization issues when deploying the method in real-world settings. Moreover, our evaluation has been limited to standard image quality metrics such as PSNR and SSIM, which, while useful, may not fully reflect clinical relevance. A more thorough assessment involving expert radiologists and task-specific evaluations is essential before considering clinical deployment.

\newpage
\section*{NeurIPS Paper Checklist}

\begin{enumerate}

\item {\bf Claims}
    \item[] Question: Do the main claims made in the abstract and introduction accurately reflect the paper's contributions and scope?
    \item[] Answer: \answerYes{} 
    \item[] Justification: We list our contributions under introduction, clearly outlining the claims we make in the paper. These are all summarized in the abstract. We also provide theorems to motivate the rules we use for noise schedule and hyperparameter selections.

\item {\bf Limitations}
    \item[] Question: Does the paper discuss the limitations of the work performed by the authors?
    \item[] Answer: \answerYes{} 
    \item[] Justification: We have addressed the limitations under Section \ref{sec:limit}.

\item {\bf Theory assumptions and proofs}
    \item[] Question: For each theoretical result, does the paper provide the full set of assumptions and a complete (and correct) proof?
    \item[] Answer: \answerYes{} 
    \item[] Justification: Please refer to Appendices B--H for all mathematical justifications for our proofs. We state the assumptions and reference the proofs in the main paper as well.

    \item {\bf Experimental result reproducibility}
    \item[] Question: Does the paper fully disclose all the information needed to reproduce the main experimental results of the paper to the extent that it affects the main claims and/or conclusions of the paper (regardless of whether the code and data are provided or not)?
    \item[] Answer: \answerYes{} 
    \item[] Justification: We explain all the dataset and experimental details in Appendix J. The code is publicly available at \url{https://github.com/MRSRL/OptDiff}.

\item {\bf Open access to data and code}
    \item[] Question: Does the paper provide open access to the data and code, with sufficient instructions to faithfully reproduce the main experimental results, as described in supplemental material?
    \item[] Answer: \answerYes{} 
    \item[] Justification: We use existing publicly available datasets, and our code is publicly available at \url{https://github.com/MRSRL/OptDiff}.

\item {\bf Experimental setting/details}
    \item[] Question: Does the paper specify all the training and test details (e.g., data splits, hyperparameters, how they were chosen, type of optimizer) necessary to understand the results?
    \item[] Answer: \answerYes{} 
    \item[] Justification: We detail data splits, how we tuned the hyperparameters for the baselines under Appendix J. For our method, the optimization and hyperparameter selection pipeline is studied in depth under Section 4, OptDiff Pipeline.

\item {\bf Experiment statistical significance}
    \item[] Question: Does the paper report error bars suitably and correctly defined or other appropriate information about the statistical significance of the experiments?
    \item[] Answer: \answerYes{} 
    \item[] Justification: We have extensive results tables that report average metrics with standard deviations for validation and test datasets. 

\item {\bf Experiments compute resources}
    \item[] Question: For each experiment, does the paper provide sufficient information on the computer resources (type of compute workers, memory, time of execution) needed to reproduce the experiments?
    \item[] Answer: \answerYes{} 
    \item[] Justification: We state the GPUs used under Appendix J and report runtimes on Fig 1 as well as tables under Appendix M.
    
\item {\bf Code of ethics}
    \item[] Question: Does the research conducted in the paper conform, in every respect, with the NeurIPS Code of Ethics \url{https://neurips.cc/public/EthicsGuidelines}?
    \item[] Answer: \answerYes{} 
    \item[] Justification: We conform with the code of ethics.

\item {\bf Broader impacts}
    \item[] Question: Does the paper discuss both potential positive societal impacts and negative societal impacts of the work performed?
    \item[] Answer: \answerYes{} 
    \item[] Justification: We discuss broader impacts under Appendix O.
    
\item {\bf Safeguards}
    \item[] Question: Does the paper describe safeguards that have been put in place for responsible release of data or models that have a high risk for misuse (e.g., pre-trained language models, image generators, or scraped datasets)?
    \item[] Answer: \answerNA{} 
    \item[] Justification: Our paper poses no such risks.

\item {\bf Licenses for existing assets}
    \item[] Question: Are the creators or original owners of assets (e.g., code, data, models), used in the paper, properly credited and are the license and terms of use explicitly mentioned and properly respected?
    \item[] Answer: \answerYes{} 
    \item[] Justification: The existing assets in this paper have been properly referenced or cited.
    \begin{itemize}
        \item The answer \answerNA{} means that the paper does not use existing assets.
        \item The authors should cite the original paper that produced the code package or dataset.
        \item The authors should state which version of the asset is used and, if possible, include a URL.
        \item The name of the license (e.g., CC-BY 4.0) should be included for each asset.
        \item For scraped data from a particular source (e.g., website), the copyright and terms of service of that source should be provided.
        \item If assets are released, the license, copyright information, and terms of use in the package should be provided. For popular datasets, \url{paperswithcode.com/datasets} has curated licenses for some datasets. Their licensing guide can help determine the license of a dataset.
        \item For existing datasets that are re-packaged, both the original license and the license of the derived asset (if it has changed) should be provided.
        \item If this information is not available online, the authors are encouraged to reach out to the asset's creators.
    \end{itemize}

\item {\bf New assets}
    \item[] Question: Are new assets introduced in the paper well documented and is the documentation provided alongside the assets?
    \item[] Answer: \answerYes{} 
    \item[] Justification: Our code is publicly available at \url{https://github.com/MRSRL/OptDiff}; the method and implementation details are described in Section 4 and Appendix J.

\item {\bf Crowdsourcing and research with human subjects}
    \item[] Question: For crowdsourcing experiments and research with human subjects, does the paper include the full text of instructions given to participants and screenshots, if applicable, as well as details about compensation (if any)? 
    \item[] Answer: \answerNA{} 
    \item[] Justification: We work with open source datasets (fastMRI, ImageNet, FFHQ).

\item {\bf Institutional review board (IRB) approvals or equivalent for research with human subjects}
    \item[] Question: Does the paper describe potential risks incurred by study participants, whether such risks were disclosed to the subjects, and whether Institutional Review Board (IRB) approvals (or an equivalent approval/review based on the requirements of your country or institution) were obtained?
    \item[] Answer: \answerNA{} 
    \item[] Justification: We work with open source datasets (fastMRI, ImageNet, FFHQ).

\item {\bf Declaration of LLM usage}
    \item[] Question: Does the paper describe the usage of LLMs if it is an important, original, or non-standard component of the core methods in this research? Note that if the LLM is used only for writing, editing, or formatting purposes and does \emph{not} impact the core methodology, scientific rigor, or originality of the research, declaration is not required.
    \item[] Answer: \answerNA{}  
    \item[] Justification: LLMs were only used for writing and editing assistance.

\end{enumerate}

\end{document}